\documentclass[12pt]{article}

\usepackage{newtxtext,newtxmath}

\usepackage{graphicx}

\usepackage[letterpaper,margin=1in]{geometry}

\renewenvironment{abstract}
	{\quotation}
	{\endquotation}

\date{}

\makeatletter
\renewcommand{\fnum@figure}{\textbf{Figure \thefigure}}
\renewcommand{\fnum@table}{\textbf{Table \thetable}}
\makeatother
\usepackage{soul}

\usepackage{scicite}
\usepackage{placeins}
\usepackage{multirow}
\usepackage{makecell}
\usepackage{url}
\usepackage{subcaption, booktabs}
\usepackage[table]{xcolor}
\definecolor{digitteal}{HTML}{2FA79A}
\definecolor{tealdark}{HTML}{1D7A70}
\definecolor{tealtint}{HTML}{E4F4F1}
\definecolor{coral}{HTML}{C0504D}
\definecolor{cream}{HTML}{F6EEE3}
\definecolor{inkgray}{HTML}{7A7A7A}

\newcommand{\best}[1]{\textbf{#1}}

\def\scititle{
    Humanoid SRL: Augmenting Humanoids with Supernumerary Limbs for Coordinated Loco-Manipulation
    
}
\title{\bfseries \boldmath \scititle}

\author{
	Fukang~Liu$^{1\dagger}$,
    Feiyang~Wu$^1$,
    Paul Barsa$^{1\ast}$,
    Jaehwi Jang$^{1\ast}$, 
    Akshat Jain$^1$,
    \\
    Zhigen Zhao$^1$,
    Yipu Chen$^1$,
    Zhaoyuan Gu$^1$,
    Nikhil Vemuri$^1$, 
    \\
    Anshul Chandaliya$^1$,
    Kaijie Zhu$^1$,  
    Sean Lublinsky$^1$, 
    Matthew Zhu$^1$,
    \\
    Tony G. Chen$^1$,
	Ye~Zhao$^{1\dagger}$\and
	\small $^{1}$Georgia Institute of Technology, Atlanta, GA, USA.\and
    \small$^\ast$These authors contributed equally to this work.\and
	\small$^\dagger$Corresponding authors. Email: fukangliu@gatech.edu, yezhao@gatech.edu\and
}

\begin{document}

\maketitle

\begin{abstract} \bfseries \boldmath

Humanoid robots are envisioned to perform diverse human-level tasks through their anthropomorphic morphology. Their dual-arm configuration enables a broad range of everyday actions. Yet, as in humans, this morphology can constrain loco-manipulation when both arms are occupied or additional contacts are required beyond their reach. 
Supernumerary robotic limbs (SRLs) offer a way to extend these capabilities, but their integration requires the added limb to be coordinated with whole-body motion and balance. In this study, we develop and integrate a supernumerary arm into a humanoid platform to investigate how morphological augmentation reshapes its feasible action space. Through quantitative analysis and hardware experiments across two configurations, we show how the additional limb expands whole-body interaction capabilities by enabling manipulation and support to be coordinated beyond the constraints of the native dual-arm configuration.
To control this augmented morphology, we leverage model-based trajectory optimization (TO) to generate dynamically feasible motions for goal-conditioned reinforcement learning (RL) controllers.
The controllers are trained in simulation and transferred to hardware in a zero-shot manner, enabling coordinated whole-body skills for contact-rich tri-arm loco-manipulation. 
We further show vision-guided autonomous execution on selected tasks using diffusion policies trained from hardware demonstrations. 
Our findings suggest that humanoid morphology need not be restricted to human-like forms, but can be reimagined as an extensible design space for robotic augmentation.

\end{abstract}

\section*{Project website}
Videos of the experiments are available at \url{https://humanoid-srl.github.io/}.

\section*{Introduction}
Humans evolved a pair of versatile forelimbs adapted for skilled interaction with tools and complex environments. This bilateral arm configuration provides an effective combination of dexterity and redundancy, supporting a wide range of bimanual actions and fine control across diverse scenarios. Yet this morphology also imposes inherent limitations on reachable workspace and the availability of arms for additional interaction. In tasks such as complex assembly and cooperative lifting, maintaining existing contacts can restrict access to other task locations, even with adaptive whole-body coordination. When both arms are occupied, additional manipulation or support may require reorganizing the task or seeking assistance.
Humanoid robots inherit this anthropomorphic dual-arm design, enabling them to operate in human-centered environments while exhibiting increasingly capable whole-body coordination across locomotion and manipulation~\cite{radosavovic2024realworld, gu2025humanoidsurvey, ze2024generalizable, teng2025robust}. Although whole-body strategies broaden the ways these robots interact with their surroundings, the availability and reach of the native arms continue to constrain their loco-manipulation capabilities. These constraints become particularly consequential when both arms are committed to an ongoing task or when additional contacts are required across spatially separated regions.

Supernumerary robotic limbs (SRLs) for humans have demonstrated how independently actuated appendages can expand the available action space~\cite{yang2021supernumerary, nguyen2019soft, vatsal2021wearable}. Users can learn to coordinate additional degrees of freedom (DoFs) alongside their natural limbs, enabling concurrent interactions beyond those supported by the native body alone~\cite{dominijanni2023human, lisini2025exploiting}. SRLs have also been explored for load sharing and physical support during manipulation, including overhead tasks and assembly, allowing the added limb to assist while the natural arms remain engaged~\cite{parietti2014supernumerary, luo2021modeling, tu2023task, parietti2016supernumerary}. 
These studies motivate morphological augmentation as a means of extending physical interaction capabilities through additional functional contacts.

Motivated by these findings, we investigate SRLs as a morphological augmentation for humanoid robots. We develop a lightweight four-DoF supernumerary arm that can be mounted on the humanoid's neck or back, with one additional arm used in each configuration. The added limb functions as a complement to the native arms, enabling new modes of multi-contact interaction and parallel manipulation. Through quantitative analysis and hardware experiments, we investigate how this augmentation reshapes the feasible action space and expands whole-body interaction capabilities. In particular, we study how loco-manipulation and support can be coordinated across the augmented body beyond the constraints of its native dual-arm configuration.
Realizing this augmentation requires more than controlling an additional manipulator. The arm's kinematic structure and attachment must provide useful access while remaining compatible with the existing body's motion. Its added mass and movement couple to whole-body dynamics, while environmental contacts may provide support or introduce disturbances. These couplings require the added limb's motion and contact forces to be coordinated with the rest of the body to maintain balance.

Model-based control methods have been fundamental to advances in humanoid whole-body behaviors~\cite {gu2025humanoidsurvey, wensing2023optimization}, enabling coordinated loco-manipulation through explicit consideration of robot dynamics and contact constraints~\cite{koenemann2015whole, khazoom2024tailoring}.
Their real-time implementation requires balancing model fidelity against computational cost. Simplified models enable more tractable online optimization~\cite{kajita2003biped, di2018dynamic}, whereas full-body~\cite{koenemann2015whole, neunert2018whole, mastalli2022agile, khazoom2024tailoring} or kino-dynamic~\cite{meduri2023biconmp,adu2023exploring} formulations capture more detailed system dynamics but place greater demands on computation and contact-state estimation. Control performance can also be affected by uncertainty in robot dynamics and deviations from the assumed contact conditions, particularly during complex physical interactions. For SRL-augmented humanoids, the additional DoFs and broader range of contact configurations can further increase these demands, especially when the native limbs and SRL must coordinate different interaction roles while maintaining whole-body balance.

Learning-based approaches offer a complementary approach to humanoid whole-body control. Reinforcement learning (RL), in particular, has shown hardware-deployable behaviors by leveraging large-scale simulation and training across variations in robot dynamics and environmental conditions~\cite{radosavovic2024realworld}. However, learning complex whole-body behaviors from scratch can require substantial interaction data and extensive reward design. 
RL-based motion imitation has been widely explored for humanoid whole-body control, using reference motions to provide structured guidance for skill acquisition~\cite{peng2018deepmimic, cheng2024expressive, sleiman2026zest, liu2024opt2skill}. A common approach retargets human motion to the robot's morphology, providing supervision for learning versatile and human-like whole-body behaviors~\cite{dugar2025learning, humanplus, luo2026sonic}. However, differences in embodiment mean that retargeted trajectories do not necessarily satisfy the robot’s kinematic and dynamic constraints, particularly during contact-rich interaction. Consequently, this embodiment gap can limit motion-tracking fidelity and the range of demonstrated behaviors that the robot can reliably reproduce. For humanoids augmented with SRLs, human demonstrations introduce a further challenge, as they do not directly specify the additional limb's motion or its coordination with native structures. Human motion data can inform the existing body's behavior, but additional motion synthesis is needed to provide complete references for the augmented morphology.

Model-based contorl and optimization has also been integrated with RL to provide physically grounded guidance for whole-body control~\cite{fuchioka2023opt,liu2024opt2skill, liu2026opt2vla, zhu2025versatile, olkin2026chasing}. In particular, optimized motion references can guide policy learning while accounting for robot dynamics and physical constraints. This integration has been explored in quadrupedal control~\cite{jenelten2024dtc, fuchioka2023opt, miller2023reinforcement} and humanoid locomotion~\cite{chaikovskaya2023benchmarking, marew2024biomechanics, li2026accelerating, olkin2026chasing}. Within this line of work, Opt2Skill~\cite{liu2024opt2skill} combines full-order-dynamics-based TO with RL for humanoid whole-body loco-manipulation. Dynamically feasible motion references improve tracking and task performance, while the associated joint-torque information provides additional supervision for contact-rich skills. This establishes a foundation for learning whole-body skills with physically consistent guidance tailored to the robot’s embodiment.

\begin{figure}
\centering
\includegraphics[width=1\textwidth]{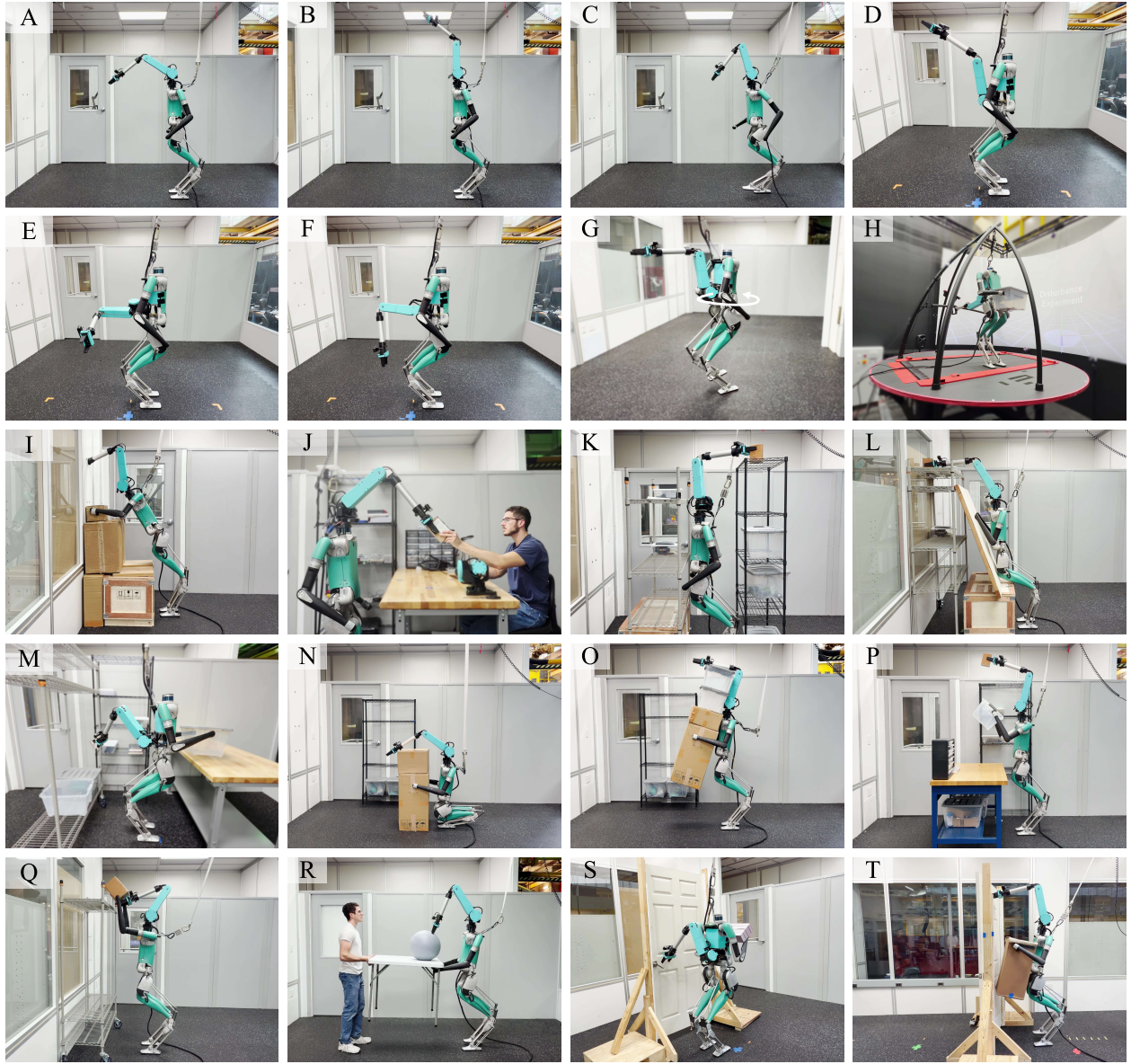}
\caption{\textbf{Configurations and real world tasks.}
The humanoid platform is equipped with a supernumerary $4$-DoF arm in neck-mounted and back-mounted embodiments. (A--G) Photographs of both configurations in distinct SRL poses. (H--T) Task snapshots illustrating the diversity of behaviors enabled by the tri-arm morphology, including reach-augmented manipulation, coordinated manipulation, dynamic loco-manipulation, and human--robot collaborative whole-body behaviors.
}
\label{fig:overview}
\end{figure}

Building on this foundation, we adapt the TO-RL framework for whole-body control of the SRL-augmented humanoid. The framework combines a TO-generated library of whole-body loco-manipulation motions with a set of goal-conditioned RL controllers. TO incorporates the augmented robot's full-body dynamics and prescribed contact constraints together with joint-torque limits, providing physically consistent supervision for coordinated motion of the native limbs and SRL. The optimized trajectories span single-arm, dual-arm and tri-arm interaction modes and guide controller training in simulation. To improve training efficiency and learn deployable controllers under partial observability, we adopt the Learn-to-Teach (L2T) privileged-learning framework~\cite{wu2025learn, miki2022learning, radosavovic2024realworld}. A teacher policy learns to track task-conditioned end-effector references using privileged simulator information, while a student policy learns to reproduce the teacher’s behavior from proprioceptive observations and the same task references. The two policies are trained concurrently, reusing simulation experience without requiring a separate student-distillation stage. The resulting student controllers are transferred to hardware in a zero-shot manner to execute coordinated whole-body behaviors with the augmented morphology.

To demonstrate that the learned whole-body controllers can also be driven by perception, we integrate a vision-based diffusion policy (DP)~\cite{chi2025diffusion} for selected autonomous tasks. The DP serves as a high-level perceptual planner that conditions on RGB-D observations and robot proprioception to predict short-horizon end-effector targets. These targets are executed by the task-specific RL controllers, allowing the augmented humanoid to perform closed-loop tri-arm behaviors from visual observations. This extension provides a vision-guided interface to the whole-body control framework without changing the underlying low-level controllers.

We evaluate the proposed system through quantitative analysis and hardware experiments on both SRL configurations (Fig.~\ref{fig:overview}).  The results show that the additional limb reshapes the humanoid's feasible action space and enables whole-body interaction beyond the constraints of the native dual-arm morphology. The TO-RL controllers transfer from simulation to hardware in a zero-shot manner and support coordinated tri-arm behaviors across manipulation, support, and dynamic whole-body interaction. We further demonstrate closed-loop autonomous execution on selected tasks using the vision-based diffusion policy.  These results establish supernumerary limbs as a viable means of extending humanoid morphology and broadening its physical interaction capabilities.

\section*{Results}

\subsection*{Morphology and Workspace Augmentation}

\begin{figure} 
\centering
\includegraphics[width=1\textwidth]{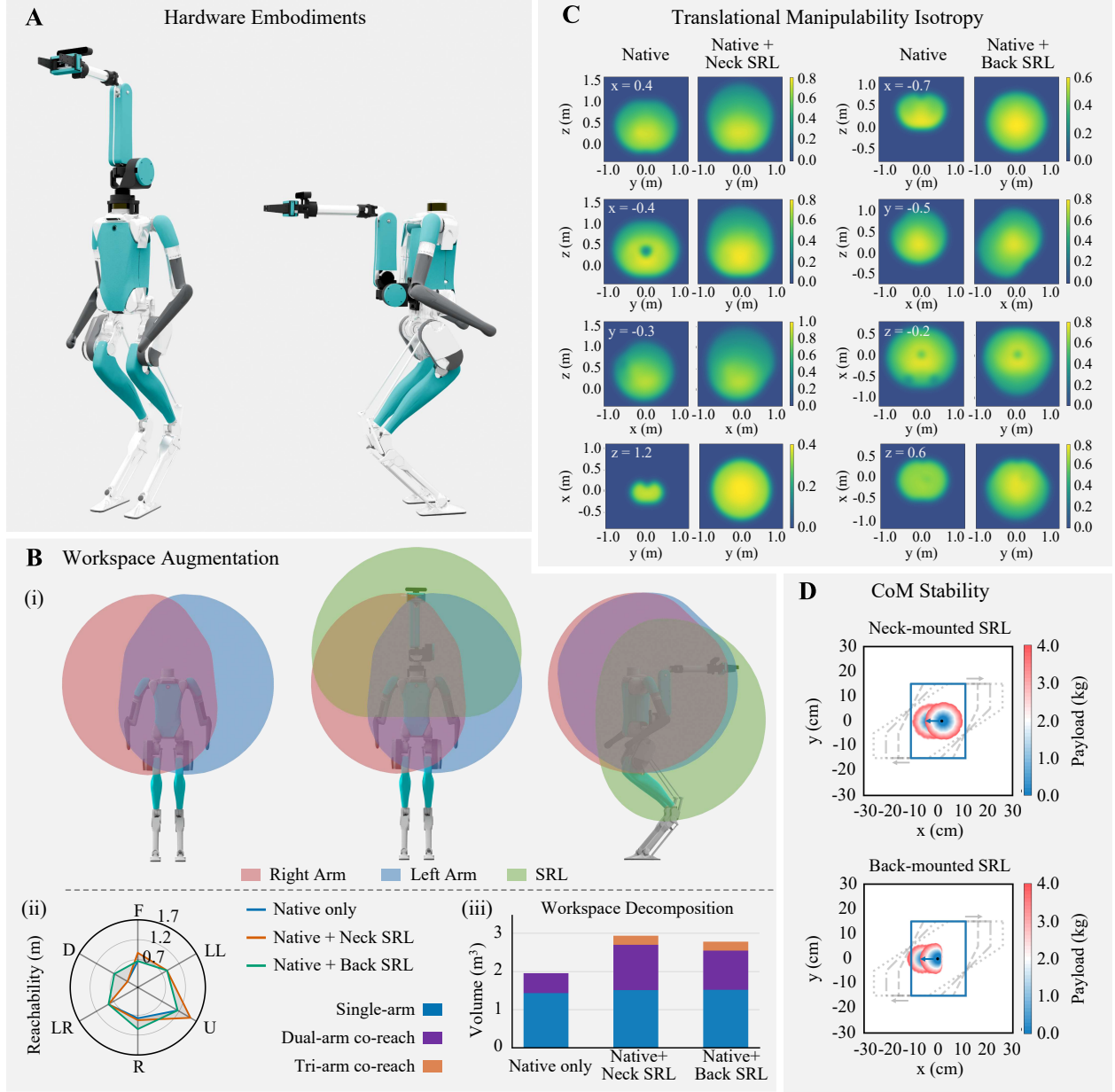}
\caption{\textbf{Morphological augmentation of workspace and support.}
(\textbf{A}) Neck-mounted and back-mounted SRL embodiments.
(\textbf{B}) Workspace augmentation in body-frame. (\textbf{i}) Reachable workspace overlays for the native arms only, native arms with neck-mounted SRL, and native arms with back-mounted SRL. (\textbf{ii}) Directional reach comparison in the body frame (F: forward; R: rearward; LL: left lateral; LR: right lateral; U: upward; D: downward). (\textbf{iii}) Workspace decomposition into single-arm volume, dual-arm co-reach, and tri-arm co-reach.
(\textbf{C}) Translational manipulability isotropy across nominal standing frame slices. The neck-mounted SRL enhances manipulability in frontal and overhead regions, whereas the back-mounted SRL improves performance in the rear, lateral, and lower regions.
(\textbf{D}) Projected CoM distributions for sampled SRL configurations and payloads of $0-4$ kg, compared with the foot-contact polygons of the selected standing configurations.
}
\label{fig:workspace}
\end{figure}

We evaluate how the added limb reshapes the humanoid's feasible action space through two mounting configurations: a neck-mounted embodiment that extends frontal and overhead reach, and a back-mounted embodiment that enlargers rearward and downward access (Fig.~\ref{fig:workspace}A; Movie S1). Both configurations use the same $4$-DoF arm module, but produce distinct augmentation patterns due to differences in attachment kinematics and whole-body coupling. These embodiments allow us to characterize how supernumerary morphology expands reachable workspace, modifies local dexterity, and influences whole-body support feasibility across spatial configurations.

Fig.~\ref{fig:workspace}B summarizes the reachable-space augmentation introduced by the added limb. The $3$D workspace overlays in Fig.~\ref{fig:workspace}B(i) show that the neck-mounted configuration preferentially expands frontal and overhead access, whereas the back-mounted configuration extends the robot's reachable set into rearward and lower regions. The directional asymmetry is quantified in Fig.~\ref{fig:workspace}B(ii). Relative to the native arms, the neck-mounted SRL increases maximum forward reach from $0.66$ m to $0.87$ m and maximum upward reach from $1.16$ m to $1.53$ m in the body frame, corresponding to $2.53$ m in the global standing pose. The back-mounted SRL increases maximum rearward reach from $0.77$ m to $1.05$ m and maximum downward reach from $0.30$ m to $0.68$ m.

Sampling-based kinematic analysis further quantifies how the added limb alters the structure of the reachable workspace. As shown in Fig.~\ref{fig:workspace}B(iii), the neck-mounted and back-mounted configurations increase total reachable volume from $1.95$ m$^3$ to $2.94$ m$^3$ ($+50.3\%$) and $2.78$ m$^3$ ($+42.3\%$), respectively. The workspace reachable by at least two arms expands from $0.52$ m$^3$ to $1.43$ m$^3$ (neck) and $1.26$ m$^3$ (back). Furthermore, Both embodiments introduce a tri-arm co-reach region of $0.24$ m$^3$. These results show that SRL augmentation expands not only total reach but also the shared workspace available to multiple arms, providing a geometric basis for coordinated multi-arm interaction.

We characterize local kinematic conditioning using translational manipulability isotropy, defined in Materials and Methods. 
Fig.~\ref{fig:workspace}C shows the best available sampled isotropy score across workspace slices expressed in the nominal standing frame. The neck-mounted SRL expands regions of favorable kinematic conditioning in the frontal and overhead workspace, whereas the back-mounted SRL extends these regions rearward and downward. In both embodiments, the added limb increases the spatial extent over which well-conditioned translational motion is available, with the direction of augmentation determined by the mounting configuration.

We finally examine whether these geometric gains are achieved without compromising static support. Fig.~\ref{fig:workspace}D shows CoM distributions for the neck-mounted and back-mounted embodiments across all reachable SRL configurations and payload from $0$ to $4$ kg. The distributions are overlaid with support polygons corresponding to a set of static standing configurations in which the feet were placed at different fore-aft and lateral offsets.
In both embodiments, the projected CoM remains within the corresponding support regions across the tested configurations, indicating that the added limb preserves static support feasibility under the loaded conditions. Detailed fixed-based tracking tests are reported in Supplementary Fig.~\ref{fig:sp_payload_tracking}. 

Together, these analyses show that the supernumerary arm reshapes the humanoid's morphological operating envelope by extending directional reach, creating new multi-arm co-reach regions, and enlarging the spatial domain available for dexterous interaction while maintaining static support feasibility. These properties provide the geometric basis for the coordinated behaviors examined in the following sections.

\begin{figure} 
\centering
\includegraphics[width=1\textwidth]{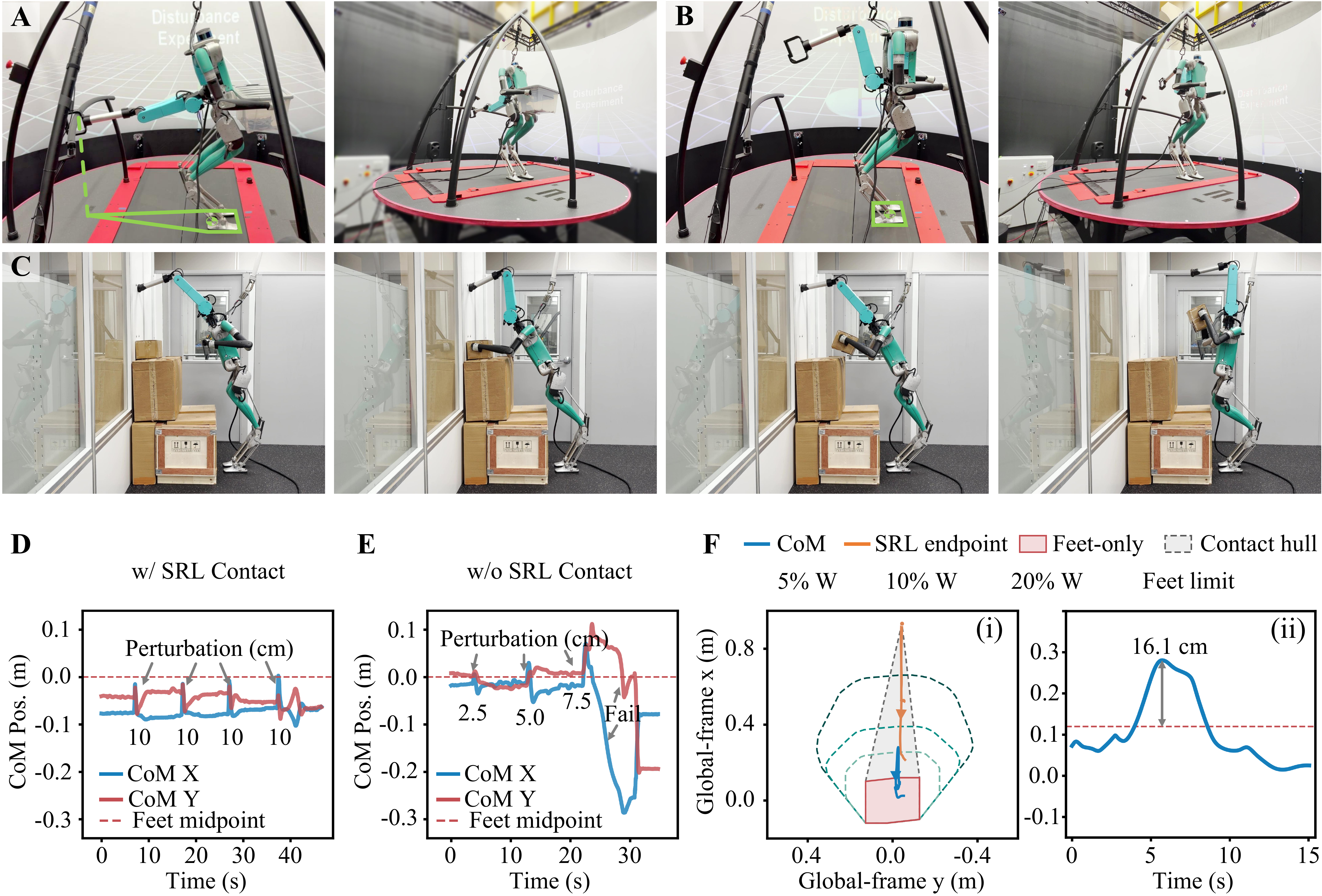}
\caption{\textbf{Support and active stability augmentation with the SRL.}
(\textbf{A, B}) CAREN perturbation experiments with and without back-mounted SRL support against the external frame.
(\textbf{C}) Forward leaning while the native arms carry a payload and the SRL presses against a wall.
(\textbf{D, E}) CoM responses with SRL support under repeated $10$ cm perturbations and without support under increasing perturbation amplitudes, including balance loss at $7.5$ cm. The $10$ cm amplitude is the experimental safety limit.
(\textbf{F}) Analysis of wall bracing.
(\textbf{i}) Global-frame CoM and SRL-endpoint trajectories relative to the feet midpoint. The feet-only support polygon and projected contact hull are shown separately from conditional static-equilibrium regions computed with unilateral contacts, wall and floor friction coefficients $\mu=1$, and wall-normal-force caps of $5\%$, $10\%$, and $20\%$ of modeled weight $W$.
(\textbf{ii}) Forward CoM position versus time, showing a peak excursion of $16.1$ cm beyond the feet-only forward boundary.
}
\label{fig:force_aug}
\end{figure}

\subsection*{Support and Stability Augmentation}
Beyond extending manipulation reach, the supernumerary arm can establish auxiliary environmental contacts that contribute to whole-body support while the native arms remain occupied. We examine this role through disturbance-rejection experiments using a Computer-Aided Rehabilitation Environment (CAREN) system and a support-assisted forward-leaning task during payload handling (Movie S2).

To assess the contribution of the additional SRL contact to disturbance tolerance, we conducted CAREN perturbation experiments with and without SRL support (Figs.~\ref{fig:force_aug}A and B). In the supported condition, the back-mounted SRL braces against the external frame, creating an additional contact that enlarges the whole-body support region beyond the feet alone. With SRL support, the robot tolerated repeated forward and backward perturbations at the maximum clip amplitude permitted by the experimental safety protocol ($10$ cm), as shown in Fig.~\ref{fig:force_aug}D. Without SRL support, the robot tolerated smaller perturbations but lost balance at $7.5$ cm when the CoM moved beyond the feet-only support polygon (Fig.~\ref{fig:force_aug}E). These results show that the auxiliary SRL contact increases disturbance tolerance by providing an additional whole-body support mode.

Beyond disturbance rejection, we investigate how SRL-assisted environmental bracing enables whole-body postures beyond feet-only support during manipulation (Fig.~\ref{fig:force_aug}C). The native arms carry a payload while the SRL braces against a wall to support the leaning posture. The global-frame trajectory in Fig.~\ref{fig:force_aug}F(ii) shows the CoM reaches a maximum forward displacement of $0.28$ m and extends $0.16$ m beyond the feet-only support boundary. Figure~\ref{fig:force_aug}F(i) shows the projected contact hull with conditional static-equilibrium regions computed assuming unilateral frictional contact ($\mu=1$ for both wall and floor). Evaluating wall-normal-force caps of $5\%$, $10\%$, and $20\%$ of the total body weight $W$, we observe that the admissible CoM region expands alongside the allowable force limit. This demonstrates that environmental bracing broadens static support through contact force capability rather than contact geometry alone.

These experiments highlight a second role of the supernumerary arm beyond manipulation: the added limb can participate directly in whole-body support. Environmental bracing expands the feasible equilibrium region, increases perturbation resistance, and enables stable postures beyond the limits of feet-only balance.

\subsection*{Triadic Coordination}
Beyond extending reach, the SRL changes how manipulation and support roles can be distributed across the humanoid body. The additional limb provides access to spatial regions beyond the native-arm workspace while allowing the native arms to maintain existing contacts or continue separate actions. We examine these capabilities through two interaction regimes: \emph{reach-augmented manipulation}, where the SRL operates in workspace regions that are difficult or inaccessible for the native arms, and \emph{coordinated augmentation}, where all three arms contribute to a shared manipulation objective.

\begin{figure} 
\centering
\includegraphics[width=1\textwidth]{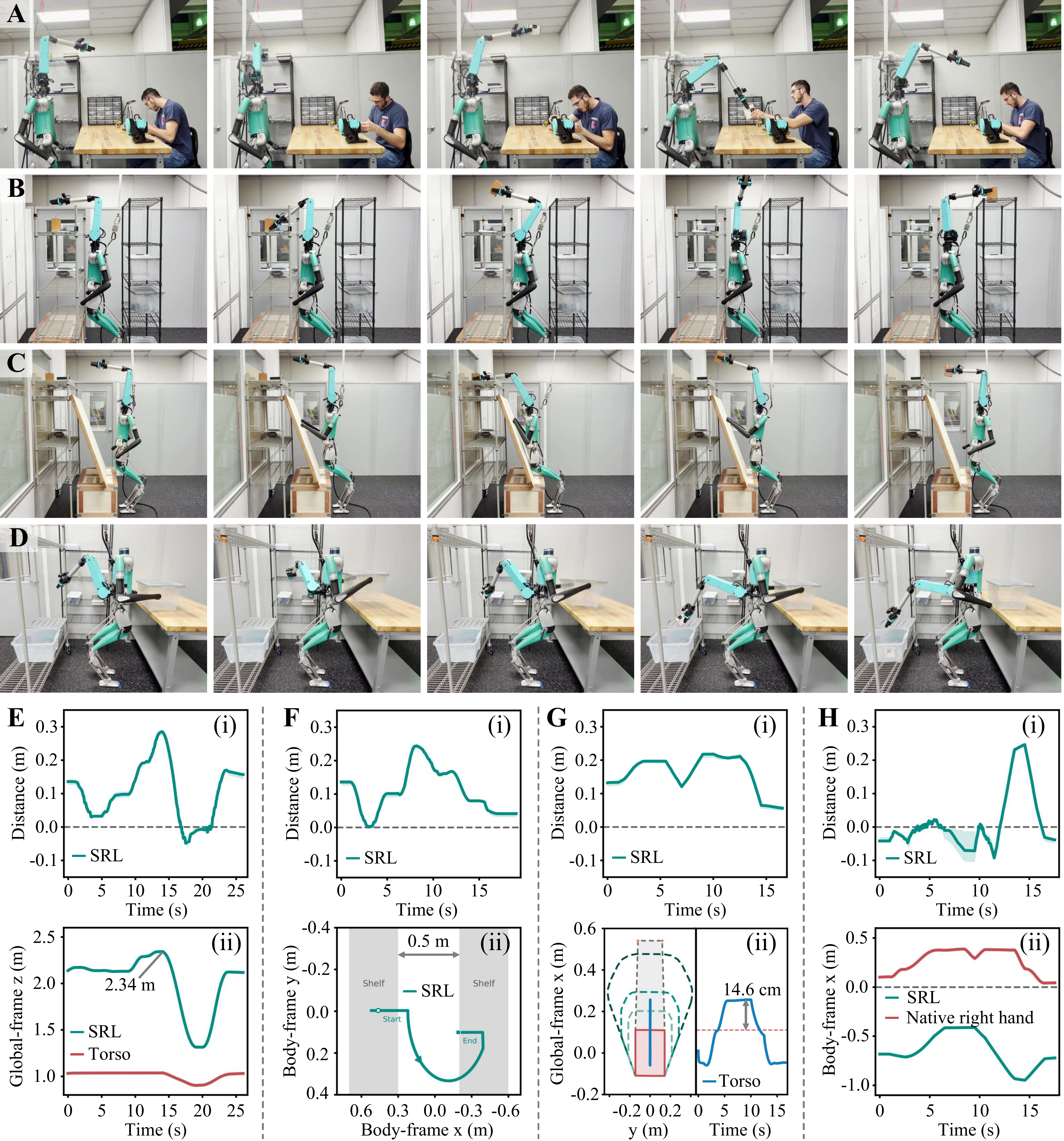}
\caption{\textbf{Reach-augmented manipulation with the SRL.}
\textbf{(A--D)} High-shelf retrieval, shelf transfer without torso reorientation, native-arm-supported reach-around, and concurrent rear/front manipulation.
\textbf{(E--H)} Corresponding reference-based analyses.
\textbf{(i)} Body-frame signed distance to the sampled native-arm workspace; positive values indicate points outside, and shading indicates voxel-size sensitivity ($15$--$25$ mm).
\textbf{(ii)} World-frame SRL and torso heights \textbf{(E)}; initial-heading-frame SRL trajectory between shelves \textbf{(F)}; projected CoM trajectory and conditional static-support regions \textbf{(G)}; and instantaneous body-frame SRL and right-hand positions \textbf{(H)}.}
\label{fig:independent_aug}
\end{figure}

\subsubsection*{Reach-Augmented Manipulation}
In reach-augmented scenarios, the SRL operates in regions outside or near the native-arm workspace, while the native arms maintain other manipulation or support roles (Movie S3). Across the four tasks, Figs.~\ref{fig:independent_aug}E--H(i) compare the SRL end-effector trajectories with the sampled native-arm workspace in a common body frame. Positive signed distance indicates that the SRL reference lies outside the sampled native-arm reachable set, whereas negative values indicate overlap. 
The shaded bands indicate sensitivity to workspace discretization.
Overall, substantial portions of the SRL references extend beyond the native-arm workspace, showing how the added limb provides access to spatial regions not readily available to the native arms.

The neck-mounted SRL retrieves an object from the top layer of a tall shelf and hands it to a human seated across the table (Fig.~\ref{fig:independent_aug}A). The  corresponding vertical trajectory in Fig.~\ref{fig:independent_aug}E(ii) reaches a maximum world-frame height of $2.34$ m and descends to $1.31$ m during the handoff. The accompanying base-height trace shows how whole-body motion contributes to the extended vertical interaction range.

We next evaluate object transfer between spatially separated shelves under constrained body motion (Fig.~\ref{fig:independent_aug}B). The neck-mounted SRL moves an object from the upper front shelf to the rear workspace while the robot maintains an approximately fixed heading. Fig.~\ref{fig:independent_aug}F(ii) shows the SRL end-effector path in a frame fixed at the initial base pose, together with schematic shelf boundaries separated by $0.50$ m. This narrow arrangement restricts body reorientation and limits the clearance available for large native-arm movements, such as raising or sweeping the arms between the shelves.  The task demonstrates how the added limb enables object transfer across spatially separated regions when both body motion and native-arm access are constrained by the environment.

In a wall-supported retrieval task, the native arms brace against a wall while the neck-mounted SRL reaches beyond the wall boundary to retrieve an object and return it to the frontal workspace (Fig.~\ref{fig:independent_aug}C). The support analysis in Fig.~\ref{fig:independent_aug}G(ii) shows that the projected CoM reaches $0.26$ m forward of the feet midpoint, exceeding the feet-only forward boundary by $0.15$ m.  Conditional static-support regions computed using the same contact assumptions as in Fig.~\ref{fig:force_aug}F illustrate how the two native-hand contacts extend the available support while the SRL performs the manipulation. This behavior highlights a division of roles in which the native arms provide environmental support and the added limb accesses the manipulation region.

Finally, Fig.~\ref{fig:independent_aug}D demonstrates concurrent manipulation across spatially separated workspaces. With the back-mounted configuration, the SRL retrieves and places an object in the rear workspace while the native arms simultaneously manipulate a separate object at a frontal tabletop. The body-frame trajectories in Fig.~\ref{fig:independent_aug}H(ii) show that the SRL remains in the rear workspace while the native right hand remains in front of the body. This demonstrates that the SRL can maintain a rearward manipulation role without displacing the native arms from their frontal task.

\subsubsection*{Coordinated Augmentation}
In coordinated augmentation scenarios, the SRL works together with the native arms toward a shared manipulation objective (Movie S4). Rather than acting independently in an extended workspace, the SRL participates through relay, object-relative manipulation, or added contact support. These tasks show how tri-arm coordination enable continuous manipulation across multiple workspace regimes and improves stability when an additional contact is required.

\begin{figure}
\centering
\includegraphics[width=1\textwidth]{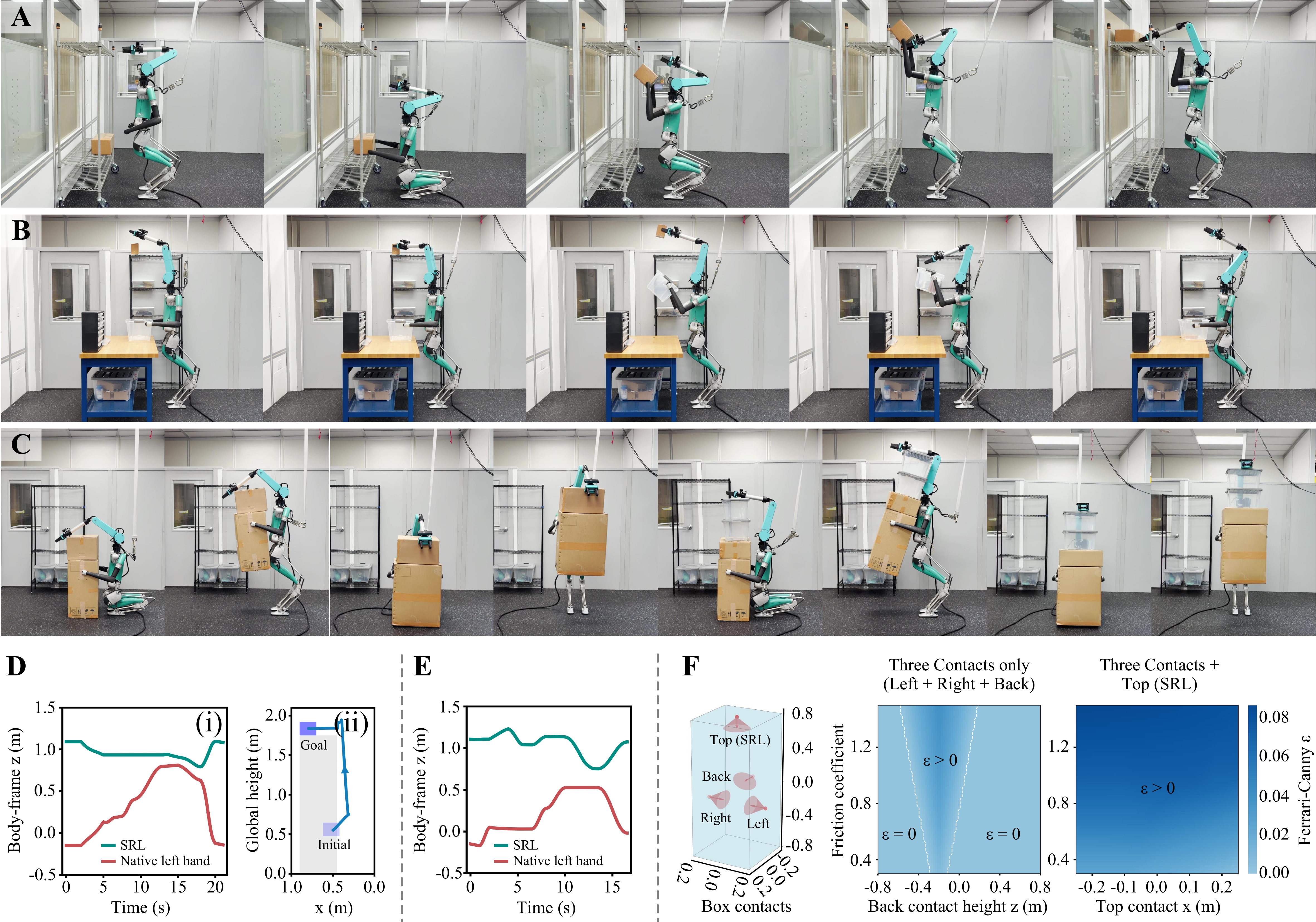}
\caption{\textbf{Coordinated augmentation through tri-arm manipulation.}
(\textbf{A}) Shelf transfer, in which the native arms lift a box from a low shelf and the SRL assists in final placement on a high shelf. 
(\textbf{B}) Coordinated loading, in which the native arms lift a container while the SRL retrieves an object from a top shelf and places it into the container. 
(\textbf{C}) Stacked-box stabilization, in which the native arms lift a multi-layer box stack while the SRL provides an additional stabilizing contact. 
(\textbf{D})(\textbf{i}) The vertical motion of the SRL and native left hand in the instantaneous body frame during during shelf transfer.
(\textbf{D})(\textbf{ii}) The full object path to the final high-shelf placement.
(\textbf{E}) The heights of the SRL and native left hand in the instantaneous body frame during coordinated loading.
(\textbf{F}) Contact schematic and Ferrari--Canny sensitivity analysis for stacked-box stabilization, showing that the added SRL contact increases wrench support and enlarges the region of positive $\epsilon$ values under a frictional contact model.
}
\label{fig:coordinated_aug}
\end{figure}

In the shelf-transfer task (Fig.~\ref{fig:coordinated_aug}A), the native arms retrieve a box from a lower shelf and lift it upward, but the final placement lies beyond their practical range for stable insertion. Once the box reaches the upper transfer region, the SRL engages and guides the final placement onto the high shelf. The vertical motion traces in Fig.~\ref{fig:coordinated_aug}D(i) show the coordinated vertical motions of the SRL and native left hand in the body frame, while Fig.~\ref{fig:coordinated_aug}D(ii) captures the full motion of the object from the initial pose to the final placement at $1.8$ m height and $0.8$ m forward relative to the torso. This task illustrates a workspace relay in which the object moves continuously from native-arm acquisition to SRL-assisted placement without intermediate drops or body repositioning.

A different coordination pattern appears in the loading task (Fig.~\ref{fig:coordinated_aug}B). The native arms first grasp and lift a container from a desk, establishing a stable support frame in front of the torso. The SRL then reaches to a top shelf, retrieves a second object, and places it into the held container. The corresponding body-frame traces in Fig.~\ref{fig:coordinated_aug}E show the coordinated vertical motions of the SRL and native left hand during this behavior. Here, the SRL does not merely extend reach; it performs object transfer relative to a structure maintained by the native arms.

The stacked-box task (Fig.~\ref{fig:coordinated_aug}C) highlights a third mode of coordination, in which the SRL contributes through an added stabilizing contact. The native arms lift a multi-layer box assembly while the SRL contacts the upper box surface to reduce tipping and improve stability during the squat-and-lift motion. Front and side views of three-layer and four-layer stacks illustrate how the additional contact supports the object though the lift. To examine the contributions of the additional contact, we numerically evaluate grasp-wrench quality across different contact locations and assumed friction coefficients using the Ferrari--Canny sensitivity analysis~\cite{ferrari1992planning}, as shown in Fig.~\ref{fig:coordinated_aug}F. The four-contact configuration shows positive $\epsilon$ values across a range of the evaluated parameters, where $\epsilon$ denotes the normalized grasp-wrench quality score, with positive values indicating force closure under the adopted frictional point-contact model. 

Across these tasks, coordinated augmentation emerges through structured role allocation among the three limbs. Depending on task phase, the native arms and SRL alternately acquire, support, guide, and stabilize the manipulated object. This coordinated use of three manipulators enables workspace relay, object-relative loading, and multi-contact stabilization that are difficult to achieve with a conventional bimanual morphology alone.

\subsection*{Dynamic and Collaborative Whole-Body Augmentation} 
The supernumerary limb extends tri-arm coordination from stationary manipulation to tasks performed during whole-body motion. In these settings, the added limb allows an ongoing manipulation or support role to be maintained while the humanoid walks, transports objects, or interacts with a human partner. We evaluate three tasks that illustrate how the SRL enables different types of interaction to proceed concurrently with locomotion (Movie S5).

\begin{figure}
\centering
\includegraphics[width=0.95\textwidth]{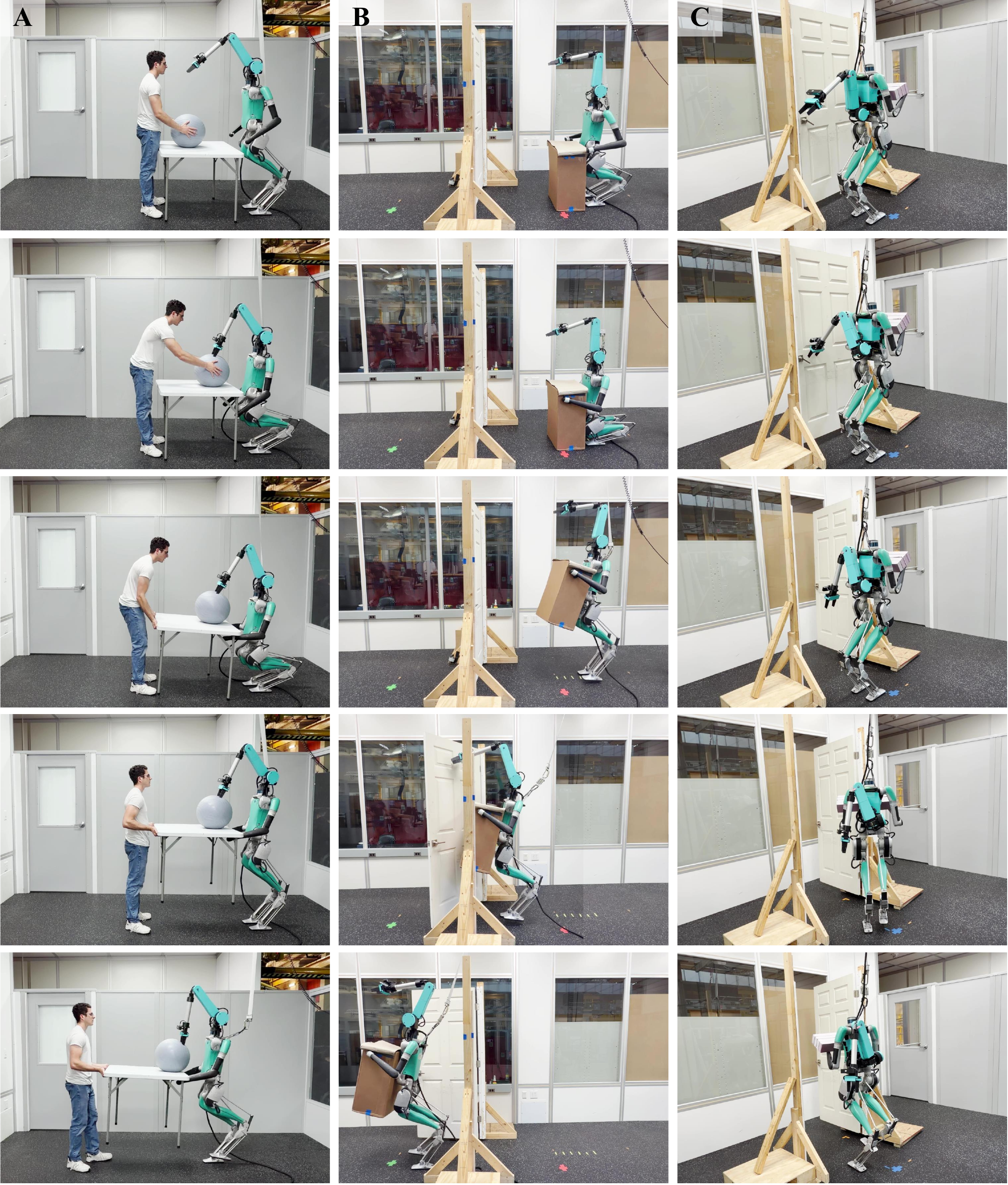}
\caption{\textbf{Dynamic and collaborative whole-body augmentation with tri-arm coordination.}
(\textbf{A}) Collaborative table transport with a human partner, where the native arms share the table load while the SRL stabilizes a loose object during lifting and walking.
(\textbf{B}) Box transport through a weighted door, with the native arms maintaining the carried object while the neck-mounted SRL pushes the door during locomotion.
(\textbf{C}) Door opening with the back-mounted SRL, which engages the handle from behind while the native arms continue carrying a box.
}
\label{fig:dynamic_aug}
\end{figure}

During collaborative table transport, the humanoid lifts and carries a table with a human partner while the SRL stabilizes an object on the tabletop (Fig.~\ref{fig:dynamic_aug}A). The native arms remain engaged with the table throughout the task, while the added limb maintains contact with the yoga ball to prevent rolling during lifting and forward walking. 
This behavior shows how the third arm can assume a secondary manipulation role without interrupting collaborative transport.

Box transport through a door illustrates how object carrying and environmental manipulation can proceed concurrently (Fig.~\ref{fig:dynamic_aug}B). The robot first squats to pick up a box and then walks forward. During transport, the neck-mounted SRL extends ahead to push a weighted door panel, allowing the robot to walk through the door without stopping. The task therefore distributes object transport and environmental manipulation across different limbs.

In a rearward door-opening task, the back-mounted SRL reaches behind the torso to engage the door handle and initiate opening while the native arms continue to hold the box (Fig.~\ref{fig:dynamic_aug}C). The added limb accesses the door before the body reorients toward the doorway, after which the humanoid proceeds through the opened passage. This configuration enables interaction from a rearward approach direction while preserving the primary transport task.

Across these scenarios, the SRL allows manipulation roles to remain active during locomotion and collaborative interaction. Rather than requiring the humanoid to complete each interaction sequentially, the additional limb enables tasks to be distributed across the body so that transport, environmental interaction, or collaboration can proceed concurrently within a coordinated whole-body behavior.

\subsection*{Vision-Guided Autonomous Control}

To demonstrate closed-loop task execution beyond predefined reference sequences, we integrate a vision-based diffusion policy (DP) as a high-level perceptual planner. The task-specific RL controllers provide feedback control for whole-body execution, but their reference targets are otherwise prescribed in advance and are not updated from visual feedback. The diffusion planner instead predicts short-horizon reference targets from updated visual and proprioceptive observations, providing a perception-driven interface to the low-level controllers.

We evaluate this extension on three tasks spanning reach-augmented manipulation, coordinated augmentation, and dynamic whole-body interaction: top-shelf retrieval and human delivery, stacked-box stabilization, and object transport during door interaction. Task-specific diffusion policies are trained from hardware demonstrations collected with the corresponding RL controllers. 

As demonstrated in Movie S6, the resulting system executes all three tasks autonomously using visual feedback to update task-level references throughout execution. These experiments show that the tri-arm whole-body controllers can be coupled with a perceptual planning layer for closed-loop autonomous operation, while the underlying whole-body execution remains handled by the same RL controllers used in the preceding experiments.

\section*{Discussion}
Our results show that equipping a humanoid with a supernumerary arm changes not only how far the robot can reach, but also the structure of its feasible whole-body interactions. The additional limb creates new workspaces of multi-arm co-reach and introduces environmental contacts that can provide extra manipulation or support capabilities. These changes allow interaction roles to be redistributed across the body, enabling the robot to maintain ongoing tasks while accessing otherwise difficult workspace regions and to sustain whole-body postures beyond feet-only support. In this sense, morphological augmentation changes both where the humanoid can interact and how different parts of the body can contribute to a task.

The two mounting configurations illustrate complementary ways in which the same SRL module can reshape this interaction space. The neck-mounted embodiment emphasizes frontal and overhead access, whereas the back-mounted embodiment extends rearward and downward interaction. These differences are reflected in the workspace, manipulability, and coordinated behaviors observed across the experiments. In our system, these morphological capabilities are realized through unified whole-body coordination rather than by treating the added limb as an independent manipulator. The TO-RL framework provides physically consistent motion guidance together with feedback control, allowing the SRL and native limbs to play different roles as task demands change. The resulting behaviors show that tri-arm coordination introduces new ways of distributing manipulation and support across the body rather than simply adding another redundant end-effector.

The vision-guided experiments provide an initial demonstration that these capabilities can also be connected to perception for closed-loop autonomous execution. Task-specific diffusion policies were evaluated on selected behaviors rather than across the full task suite, and broader generalization to new environments and objects, as well as longer-horizon interactions, remains to be studied. The present hardware study also considers two mounting configurations of a single SRL design rather than a systematic optimization over possible morphologies. 
In addition, the grasp-support analysis relies on simplified frictional contact assumptions. Future work could jointly optimize limb geometry and attachment for different task families. 
Extending this approach to multiple SRLs would require selecting their optimal number and arrangement to broaden functional capability while respecting the humanoid's payload limits and avoiding self-collision. Richer contact estimation could further support whole-body interaction, alongside advances in perception-driven control for less structured environments and longer-horizon human--robot collaboration.

Taken together, these findings support a broader view of humanoid robots as extensible morphologies rather than fixed replicas of the human body. A supernumerary limb need not serve only as additional reach or payload capacity; it can change how physical interaction roles are distributed across the robot and its environment. This perspective highlights a design and control space for humanoids in which morphological augmentation and whole-body coordination jointly expand physical interaction capabilities.

\section*{Materials and Methods}
\subsection*{Humanoid and SRL Hardware}
\paragraph*{Humanoid platform.}
Experiments were conducted using a Digit humanoid robot from Agility Robotics. This robot has a $30$-DoF platform with $20$ actuated joints, including two $4$-DoF arms and two $6$-DoF leg modules. The ankle and shin joints incorporate passive compliance using leaf-spring four-bar linkages. Toe pitch and roll are actuated indirectly via pushrod-coupled dual motors. The platform weighs approximately $48$ kg, and is designed for dynamic locomotion and mobile manipulation. Custom communication and control interfaces were implemented to enable synchronized multi-arm coordination and real-time integration of SRL-mounted perception.
\paragraph*{SRL mechanical design and actuation.}
To enable morphological augmentation, we designed a modular $4$-DoF supernumerary robotic limb (SRL) that can be mounted either at the neck or at the back of the humanoid. The SRL consists of a $3$-DoF shoulder assembly (roll, pitch, yaw), a remotely actuated elbow joint driven by a $3$:$1$ belt transmission, and a wrist interface for interchangeable end-effectors. The arm is designed for low distal inertia through a gimbal-style shoulder structure and by relocating the elbow drive to the joint base using remote actuation. The SRL has a total length of $83.0$ cm  and a total mass of $4.3$ kg, including drivers, structural linkage, and wiring. We used a compact rack-and-pinion parallel gripper for loco-manipulation tasks. The end-effector supports up to $2$ kg continuous payload and $4$ kg short-duration peak payload. 

The primary elbow link ($30.8$ cm) was machined from Al-$6061$ to withstand transmission loads, whereas the wrist assembly used a high-strength Al$7075$ shaft coupled to the drive pulley through a keyed interface. Structural adapters, including the neck and back mounting brackets, were fabricated in PLA. Shoulder roll, yaw, and elbow joints were actuated using RobStride RS-$02$ motors, while the shoulder pitch used a RS-$03$ motor. Actuator selection was guided by inverse dynamics simulation to ensure torque-speed feasibility over representative SRL trajectories. Additional implementation details, including battery and communication architecture, are provided in Supplementary Materials.

\paragraph*{Manipulability analysis.}
We characterize local kinematic conditioning using translational manipulability isotropy~\cite{yoshikawa1985manipulability}. Let
$\mathbf{q}\in\mathbb{R}^{n_q}$ denote the robot configuration and $\mathbf{v}\in\mathbb{R}^{n_v}$ its generalized velocity. The configuration contains the floating-base
position and unit-quaternion orientation together with the modeled internal joint coordinates.
For arm $i\in\{\mathrm{L, R, SRL}\}$, let $\mathbf{J}_i(\mathbf{q})\in\mathbb{R}^{3\times n_v}$ denote the world-frame translational end-effector Jacobian, including the six floating-base velocity columns. We define $\mathbf{v}=\mathbf{D}\nu$, where $\nu\in\mathbb{R}^{n_v}$ is dimensionless and $\mathbf{D}=\operatorname{diag}(s_1,\ldots,s_{n_v})\in\mathbb{R}^{n_v\times n_v}$ specifies reference velocity scales. We use $1$ m/s for base translation and $1$ rad/s for base rotation and internal rotational coordinates; these fixed normalization conventions reproduce the unweighted SI-coordinate calculation and are not actuator speed limits. We define the velocity-scaled Jacobian as 
\begin{equation}
    \bar{\mathbf{J}}_i(\mathbf{q})
    = \mathbf{J}_i(\mathbf{q})\mathbf{D}
    \in\mathbb{R}^{3\times n_v},
\end{equation}
which maps normalized generalized velocities to end-effector translational velocities. The isotropy score is
\begin{equation}
I_i(\mathbf{q}) =
\left(
\frac{\sigma_{\min}\left(\bar{\mathbf{J}}_i(\mathbf{q})\right)}
{\sigma_{\max}\left(\bar{\mathbf{J}}_i(\mathbf{q})\right)}
\right)^2,
\end{equation}
where $\sigma_{\min}$ and $\sigma_{\max}$ denote the smallest and largest singular values, respectively. Thus $I_i\in[0,1]$, with larger values indicating more uniform local translational velocity capability under the stated scaling. 

During workspace sampling, the base pose and non-evaluated joint positions were held at a common nominal configuration, while the evaluated arm was varied within its joint limits. The floating-base velocity columns were retained in the Jacobian. The resulting isotropy score describes unconstrained kinematic conditioning without enforcing planted-foot or linkage-loop velocity constraints. For each spatial cell $V$, $I_i(V)$ was defined as the highest isotropy score among accepted samples of arm i whose end-effector positions lay within that cell. The native-arm and augmented maps were computed as $\max\{I_{\mathrm{L}}(V),I_{\mathrm{R}}(V)\}$ and $\max\{I_{\mathrm{L}}(V),I_{\mathrm{R}}(V),I_{\mathrm{SRL}}(V)\}$, respectively. Cells without accepted samples were assigned zero for visualization.

\subsection*{Reinforcement Learning-based Whole-Body Control}
\paragraph*{Problem formulation.}
We formulate the robot whole-body control problem as a partially observable Markov Decision Process (POMDP). At each time step $t$, the robot receives an observation $o_t$ generated from the underlying state $s_t \in \mathcal{S}$ according to $o_t\sim \mathcal{O}(\cdot|s_t)$. The policy selects an action $a_t\in \mathcal{A}$ according to $a_t \sim\pi(\cdot | o_t)$, after which the system evolves according to the transition dynamics $s_{t+1} \sim p(\cdot|s_t,a_t)$, and receives an instantaneous reward $r(s_t, a_t) \in \mathbb{R}$. The objective is to learn a policy that maximizes the expected discounted return,
\begin{equation}
\pi^\ast =
\arg\max_{\pi}
\mathbb{E}_{\rho_0,p,\mathcal{O},\pi}
\left[
\sum_{t=0}^{\infty}
\gamma^t r(s_t,a_t)
\right],
\end{equation}
\noindent where $s_0 \sim \rho_0$ and $\gamma\in [0,1)$ is the discount factor. In deployment, the observation $o_t$ includes proprioceptive measurements, recent action history, and task-conditioned reference targets, without access to privileged simulator states.

\begin{figure} 
\centering
\includegraphics[width=1\textwidth]{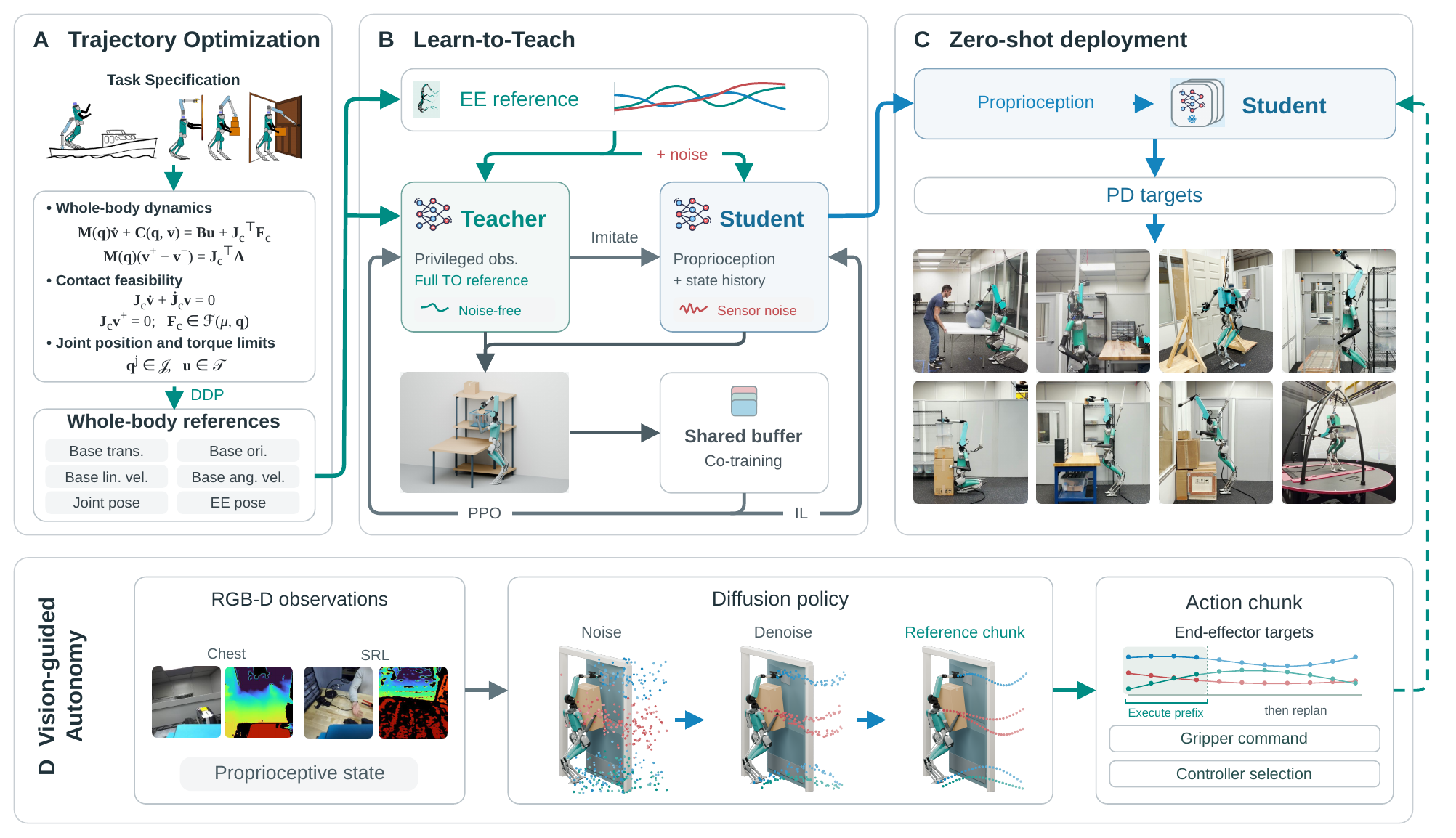}
\caption{\textbf{Hierarchical framework for supernumerary humanoid loco-manipulation.}
Trajectory optimization generates dynamically feasible whole-body references that coordinate the native limbs and SRL under contact and actuation constraints. Learn-to-Teach combines privileged teacher supervision and shared experience to train a student policy for reference tracking using noisy proprioceptive observations. The student policy transfers zero-shot to hardware, enabling coordinated manipulation, support, and locomotion through whole-body feedback control. Vision-guided references extend the same control framework to autonomous task execution.
}
\label{fig:framework}
\end{figure}

\paragraph*{RL control policy with Learn-to-Teach.}
To learn deployable whole-body controllers from the TO references, we adopt the Learn-to-Teach (L2T) privileged-learning framework~\cite{wu2025learn}. L2T jointly trains a privileged teacher policy $\pi^{\mathrm{T}}$ and a deployable student policy $\pi^{\mathrm{S}}$, with a single training stage. 
The teacher receives $\mathbf{o}^{\mathrm{T}}_t=[\mathbf{o}^{\mathrm{S}}_t,\mathbf{o}^{\mathrm{priv}}_t]$ including the student observation $\mathbf{o}^{\mathrm{S}}_t \in \mathbb{R}^{d_\mathrm{S}}$, and privileged information $\mathbf{o}^{\mathrm{priv}}_t \in \mathbb{R}^{d_\mathrm{priv}}$; whereas the student has access only to observations $\mathbf{o}^{\mathrm{S}}_t$ that is available on hardware. 
Both policies output desired joint-position targets $\mathbf{a}_t \in \mathbb{R}^{n_u}$, which are converted to joint torques through a PD controller, as shown in Fig.~\ref{fig:framework}. We train the RL policies in the MuJoCo simulator~\cite{todorov2012mujoco, ZakkaK-RSS-25}.

A key component of L2T is the joint collection of teacher- and student-generated experience. 
At each timestep, the action applied to the environment is sampled according to
\begin{equation}
\label{eq:l2t_sampling}
\mathbf{a}_t \sim
\begin{cases}
\pi^{\mathrm{S}}(\cdot \mid \mathbf{o}^{\mathrm{S}}_t),
& \text{with probability } \alpha_{\mathrm{mix}}, \\[2mm]
\pi^{\mathrm{T}}(\cdot \mid \mathbf{o}^{\mathrm{T}}_t),
& \text{with probability } 1-\alpha_{\mathrm{mix}},
\end{cases}
\end{equation}
where $\alpha_{\mathrm{mix}}$ is progressively increased during training. 
This allows teacher-generated experience to be reused for student learning while gradually exposing the policies to states induced by the deployable student. 
The overall objective combines reinforcement learning for the teacher with imitation learning for the student:
\begin{equation} 
\label{eq:l2t_loss}
\begin{aligned}
\mathcal{L}_{\mathrm{l2t}} = \mathcal{L}_{\mathrm{RL}}^{\mathrm{T}} + \lambda_{\mathrm{IL}} \, \mathcal{L}_{\mathrm{IL}}^{\mathrm{S}},
\end{aligned}
\end{equation} 
where $\lambda_{\mathrm{IL}}$ controls the contribution of the student imitation objective.
The teacher is optimized using proximal policy optimization (PPO)~\cite{schulman2017proximal}, with the standard clipped surrogate objective
\begin{equation}
\label{eq:ppo_loss}
\mathcal{L}_{\mathrm{RL}}^{\mathrm{T}}
=
-\mathbb{E}_{t}
\left[
\min
\left(
r_t(\theta)\hat{A}_t,\,
\mathrm{clip}
\left(
r_t(\theta),
1-\epsilon_\mathrm{clip},
1+\epsilon_\mathrm{clip}
\right)
\hat{A}_t
\right)
\right],
\end{equation}
where $r_{t}(\theta)$ is the probability ratio between the current and previous teacher policies, and $\hat{A}_{t}$ is the estimate advantage  (GAE).
In parallel, the student learns to imitate the teacher under its restricted observation space using
\begin{equation}
\label{eq:student_loss}
\mathcal{L}_{\mathrm{IL}}^{\mathrm{S}}
=
\mathbb{E}_{t}
\left[
\left\|
\mathbf{a}^{\mathrm{S}}_t
-
\mathbf{a}^{\mathrm{T}}_t
\right\|_2^2
\right],
\end{equation}
where
$\mathbf{a}^{\mathrm{S}}_t\sim\pi^{\mathrm{S}}(\cdot \mid\mathbf{o}^{\mathrm{S}}_t)$
and
$\mathbf{a}^{\mathrm{T}}_t=\pi^{\mathrm{T}}(\cdot\mid\mathbf{o}^{\mathrm{T}}_t)$.

Teacher and student policies are updated concurrently from the collected experience, rather than training the student in a separate distillation stage. 
This concurrent training framework improves training efficiency, while the mixed rollouts expose the teacher–student pair to the state distribution induced by the deployable policy, reducing the \textit{imitation-gap} \cite{wu2025learn}, where conventional teacher-student learning, or privileged learning framework fails to address. 
We show that with L2T, reference tracking error can be reduced up to $35\%$ with the same training budget. 
At deployment, only the student policy is retained, using proprioceptive measurements, action history, and task-conditioned reference targets without privileged simulator information. 
Details of the observation spaces, rollout schedule, network architecture, PPO implementation, and training hyperparameters are provided in the Supplementary Materials.

\subsection*{Whole-Body Trajectory Optimization}
We generated dynamically feasible reference motions using full-body trajectory optimization (TO) on the SRL-augmented humanoid model. The floating-base dynamics are governed by
\begin{equation} 
\label{eq:fbEoM}
\begin{aligned}
\mathbf{M}(\mathbf{q}) 
\dot{\mathbf{v}} + 
\mathbf{C}
(\mathbf{q, v}) = 
{\mathbf{Bu}} + 
\mathbf{J_{\rm c}^{\top}(\mathbf{q})} 
\mathbf{F_{\rm c}}.
\end{aligned}
\end{equation}
where $\mathbf{M}(\mathbf{q})\in\mathbb{R}^{n_v\times n_v}$ is the generalized mass matrix, $\mathbf{C}
(\mathbf{q, v})\in\mathbb{R}^{n_v}$ collects Coriolis, centrifugal, and gravitational terms.  ${\mathbf{B}}\in\mathbb{R}^{n_v\times n_u}$ is the actuation matrix, and $\mathbf{u}\in\mathbb{R}^{n_u}$ denotes joint-torque vector. Here, $\mathrm{n}_u$ is the number of independent torque inputs. For the active contact set, $\mathbf{J}_{\mathrm{c}}(\mathbf{q})\in\mathbb{R}^{m_c\times n_v}$ maps generalized velocities to contact velocities, and $\mathbf{F}_{\mathrm{c}}\in\mathbb{R}^{m_c}$ contains the corresponding contact-force and moment components. 

Given this whole-body dynamics, for each task, we solved a finite-horizon optimal-control problem with a prescribed contact schedule:
\begin{subequations} \label{eq:whole_to}
\begin{align} 
\label{eq:cost}
& 
\underset{
\mathbf{x},
\mathbf{u}
}
{\min} 
& &
\sum_{k=0}^{N - 1} 
\Bigl( 
\| 
\mathbf{y}[k] - 
\hat{\mathbf{y}}[k] \|^2 _\mathbf{Q} + 
\| 
\mathbf{u}[k] \|^2_{\mathbf{R}} 
\Big) + \\
&
&& 
\| 
\mathbf{y}[N] - 
\hat{\mathbf{y}}[N] \|^2 _{\mathbf{Q}_{\rm f}} \nonumber \\
\label{eq:init}
&
\text{subject to} 
&&
\mathbf{x}[0]=\mathbf{x}_\mathrm{init}, 
\mathbf{x}[k+1]=f_k(\mathbf{x}[k], \mathbf{u}[k]), \\
\label{eq:hybrid_dynamics}
&
(\textit{\rm Dynamics})         
&& 
\begin{cases} 
\mathbf{M}(\mathbf{q}) 
\dot{\mathbf{v}} + 
\mathbf{C}(\mathbf{q}, \mathbf{v})  = 
\mathbf{B} 
\mathbf{u} + 
\mathbf{J_{\rm c}^{\top}(\mathbf{q})} 
\mathbf{F_{\rm c}}\\
\mathbf{M}(\mathbf{q})\mathbf{v}^+ - 
\mathbf{M}(\mathbf{q})\mathbf{v}^- = 
\mathbf{J_{\rm c}^{\top}(\mathbf{q})} 
\boldsymbol 
\Lambda,
\end{cases}\\
\label{eq:contact}
&
(\textit{\rm Contact}) 
&& 
\begin{cases}
\mathbf{J}_{\rm c}(\mathbf{q}) 
\dot{\mathbf{v}}+
\dot{\mathbf{J}}_{\rm c}(\mathbf{q},\mathbf{v})
\mathbf{v} = 
0,\\
\textcolor{black}{
\mathbf{J}_{\rm c} (\mathbf{q})
\mathbf{v}^+ = 
0
}
\end{cases}\\
\label{eq:joint_torque_limits}
&
(\textit{\rm Limits}) 
&& 
\mathbf{q}^{\rm j} 
\in \mathcal{J}, 
\mathbf{u} \in \mathcal{T},\\
&
\text{(Friction)} 
&&
\mathbf{F}_{\rm c} 
\in \mathcal{F}(\mu, \mathbf{q}).
\hspace{1cm} 
\label{eq:friction_cone}
\end{align}
\end{subequations}
where $\mathbf{x} = [\mathbf{q};\mathbf{v}]
\in\mathbb{R}^{n_q+n_v}$ denotes the stored full-body state, and $N$ is the number of time intervals. The task-space variables $\mathbf{y} = \Phi(\mathbf{q})\in\mathbb{R}^{n_y}$ are optimized toward task-dependent targets $\hat{\mathbf{y}}$. The symmetric positive-semidefinite matrices
$\mathbf{Q},\mathbf{Q}_{\mathrm{f}}\in\mathbb{R}^{n_y\times n_y}$ and
$\mathbf{R}\in\mathbb{R}^{n_u\times n_u}$ weight tracking error and control effort, with $\|\mathbf{e}\|^2_\mathbf{Q}=\mathbf{e}^\top\mathbf{Q}\mathbf{e}$. The first rows of the dynamics and contact equations apply during continuous contact phases; the second rows describe perfectly inelastic impacts. Superscripts \(-\) and \(+\) denote pre- and post-impact velocities, respectively, while \(\boldsymbol\Lambda\) is the contact impulse and the configuration remains unchanged across impact. Contact forces and impulses are obtained from the corresponding constrained dynamics, with zero wrench assigned to inactive contacts. The sets \(\mathcal J\), \(\mathcal T\), and \(\mathcal F(\mu,\mathbf q)\) specify bounds on the internal joint coordinates \(\mathbf q^{\mathrm j}\), joint torques, and admissible contact wrenches, respectively, where \(\mu\) is the friction coefficient. Numerical settings and implementation details are provided in the Supplementary Materials.

Following the TO pipeline in Opt2Skill~\cite{liu2024opt2skill}, we solve the whole-body TO problems using DDP in Crocoddyl with rigid-body computations from Pinocchio~\cite{mastalli_crocoddyl_2020, carpentier2019pinocchio}. 
Time-indexed end-effector targets were prescribed for precise motion tracking, whereas manipulation behaviors such as box pickup were specified through sparse subgoals. Task-specific costs and contact schedules were predefined. The optimized base and joint states, together with joint torques and end-effector motions, formed the offline reference library for downstream RL policy training.

\subsection*{Vision-guided autonomous control}

To enable closed-loop autonomy for selected hardware tasks, we use a diffusion policy (DP) \cite{chi2025diffusion} as a high-level perceptual planner. The policy conditions on current robot proprioception together with RGB-D observations from the torso-mounted egocentric and SRL end-effector cameras, and predicts short horizon task-conditioned references for the task-specific RL student controllers.

Training data were collected on hardware by executing predefined reference motions with the trained student controllers while recording synchronized visual observations, robot states, and reference actions. The synchronized demonstrations were subsampled to $50$ Hz to construct the dataset
$\mathcal{D} = \left\{ (O_t, A_t)\right\}_{t=1}^N$,
where $N$ is the number of training samples. We use a single observation timestep with
$O_t = [I_t^{\mathrm{chest}}, I_t^{\mathrm{SRL}}, \mathbf{q}_t^{\mathrm{motor}}]\in\mathbb{R}^{d_o}$, which includes RGB-D observations $I^{\mathrm{chest}}$ and $I^{\mathrm{SRL}}$ from the torso egocentric and SRL cameras, respectively, together with observed motor-joint positions $\mathbf{q}^{\mathrm{motor}}\in\mathbb{R}^{n_m}$. The corresponding future action chunk is $A_t = \{[\mathbf{a}_i^{\mathrm{ref}}, a_i^{\mathrm{gripper}}, a_i^{\mathrm{mode}}]\}_{i=t}^{t+l_p-1}$, where $l_p$ is the prediction horizon, of which the first $l_a$ actions are executed before re-planning. The vector $\mathbf{a}^{\mathrm{ref}}\in\mathbb{R}^{15}$ contains the body-frame reference $3$D positions of the $5$ controlled end-effectors. The scalar commands $a_i^{\mathrm{gripper}}$ and $a_i^{\mathrm{mode}}$ specify the gripper state and encode the selected RL controller, respectively.

We model the conditional action distribution $p_\theta(A_t \mid O_t)$ using a denoising diffusion probabilistic model (DDPM) \cite{ho_denoising_2020}. At diffusion step $k$, the action chunk is perturbed as
\begin{equation}
A_t^{k} = \sqrt{\bar{\alpha}_k}\, A_t + \sqrt{1 - \bar{\alpha}_k}\, \boldsymbol{\xi}
\qquad
\boldsymbol{\xi}\sim\mathcal{N}(0,I),
\end{equation}
where $\bar{\alpha}_k$ is the cumulative noise-schedule coefficient at diffusion step $k$. The denoising network $\epsilon_\theta$, parameterized by $\theta$, is trained to recover the injected noise using
\begin{equation}
    \mathcal{L}_{\text{diff}}(\theta) 
    = \mathbb{E}_{(A_t, O_t)\sim\mathcal{D},\, k,\, \boldsymbol{\xi}}\Big[
        w(k)\,\|\boldsymbol{\xi} - \boldsymbol{\xi}_\theta(A_t^{k}, O_t, k)\|_2^2
    \Big],
\end{equation}
where $w(k)$ is a diffusion-step-dependent weighting function (Supplementary Materials), $k\sim \mathrm{Unif}\{1,\dots,K\}$, and noise $\boldsymbol{\xi}$ is sampled from a standard Gaussian distribution.

During deployment, the diffusion policy operates in a receding-horizon manner, repeatedly generating short-horizon references from updated visual and proprioceptive observations. The predicted references are linearly interpolated to match the control frequency of the selected RL student controller. The selected RL student controller, conditioned on the interpolated references and its own proprioception, generates the whole-body motion. Additional details are provided in the Supplementary Materials.


\clearpage 

%
\bibliography{humanoid} 
\bibliographystyle{sciencemag}

%
%
%
%
%
%


\section*{Acknowledgments}
The authors would like to thank Prof. Maegan Tucker for kindly providing access to her laboratory space and facilities for our hardware experiments.

\paragraph*{Funding:}
This work was supported in part by the Office of Naval Research (ONR) under Grant N000142312223, in part by the National Science Foundation (NSF) under Grant CMMI-2144309, Grant CMMI-2328254,  Grant CMMI-2611346, and in part by the United States Department of Agriculture (USDA) under Grant 2022-11065.
\paragraph*{Author contributions:}
F.L. conceived and led the project, developed the overall framework, trajectory optimization, and whole-body control, led system integration, sim-to-real transfer, hardware experiments, data collection and analysis, contributed to the SRL design, and led the manuscript writing. F.W. integrated the L2T framework, assisted with hardware experiments, and contributed to manuscript writing. P.B., A.J., N.V., A.C., S.L., and M.Z. contributed to SRL hardware design, prototyping, integration, and testing, with P.B. leading the hardware development. J.J. contributed to hardware experiments and the diffusion-policy/RL-control integration. Z.Z. and Y.C. developed and trained the diffusion policies. J.J., Z.Z. and Y.C. also contributed to manuscript writing. K.Z. assisted with humanoid-SRL hardware debugging and electrical integration. Z.G. contributed to the CAREN experiments. T.G.C. contributed technical discussions. Y.Z. provided leadership, guidance, and funding.
\paragraph*{Competing interests:}
There are no competing interests to declare.
\paragraph*{Data and materials availability:} 
All (other) data needed to evaluate the conclusions in the paper are present in the paper or the Supplementary Materials. 
\subsection*{Supplementary materials}
Supplementary Methods\\ 
Figures S1 to S4\\
Table S1 to S11\\
Movies S1 to S6\\


\newpage


\renewcommand{\thefigure}{S\arabic{figure}}
\renewcommand{\thetable}{S\arabic{table}}
\renewcommand{\theequation}{S\arabic{equation}}
\renewcommand{\thepage}{S\arabic{page}}
\setcounter{figure}{0}
\setcounter{table}{0}
\setcounter{equation}{0}
\setcounter{page}{1} 


\begin{center}
\section*{Supplementary Materials for\\ \scititle}

Fukang~Liu \textit{et al.}\\
\small Corresponding authors: Fukang Liu, fukangliu@gatech.edu; Ye Zhao, yezhao@gatech.edu
\end{center}

\subsubsection*{This PDF file includes:}
Supplementary Methods \\
Figures S1 to S4\\
Tables S1 to S11\\
Captions for Movies S1 to S6\\

\subsubsection*{Other Supplementary Materials for this manuscript:}
Movies S1 to S6\\

\newpage

\section*{Supplementary Methods}
\subsection*{SRL Hardware Specifications}

\begin{figure}[!htbp]
    \centering
    \includegraphics[width=1\linewidth]{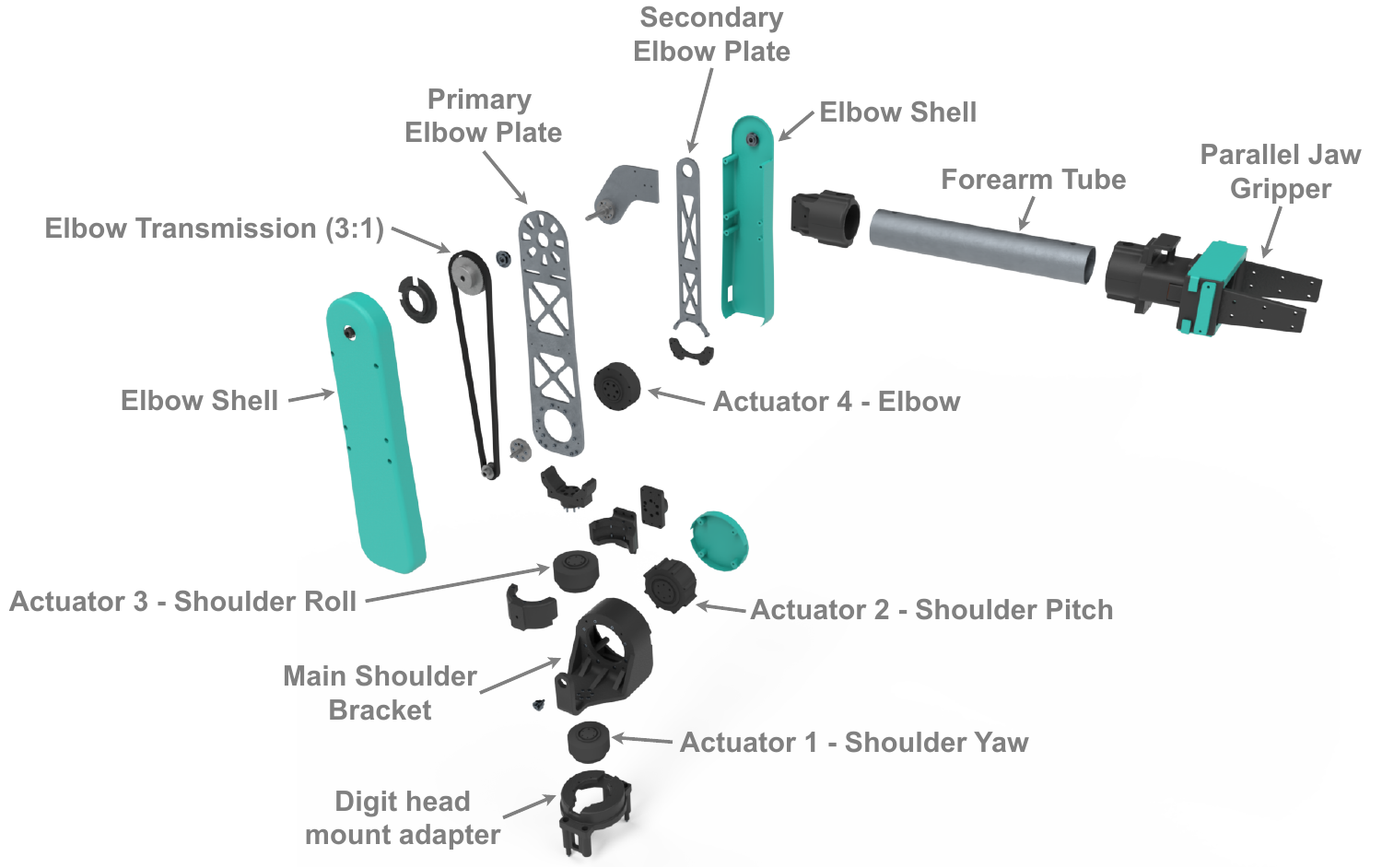}
    \caption{\textbf{Exploded view of the supernumerary arm.}. The shoulder comprises three co-located joints, with the actuators housed in high-infill 3D-printed brackets. The elbow link consists of two parallel aluminum plates that provide structural rigidity and accommodate a dual-supported bearing arrangement and belt-tensioning mechanism. The elbow is actuated remotely through a $3{:}1$ belt transmission. The hollow aluminum forearm provides structural stiffness and internal cable routing, with an interface for different end-effectors. The assembly is shown with a parallel-jaw gripper.}
    \label{fig:sr_exploded_view}
\end{figure}

\begin{figure}[!htbp]
    \centering
    \includegraphics[width=1\linewidth]{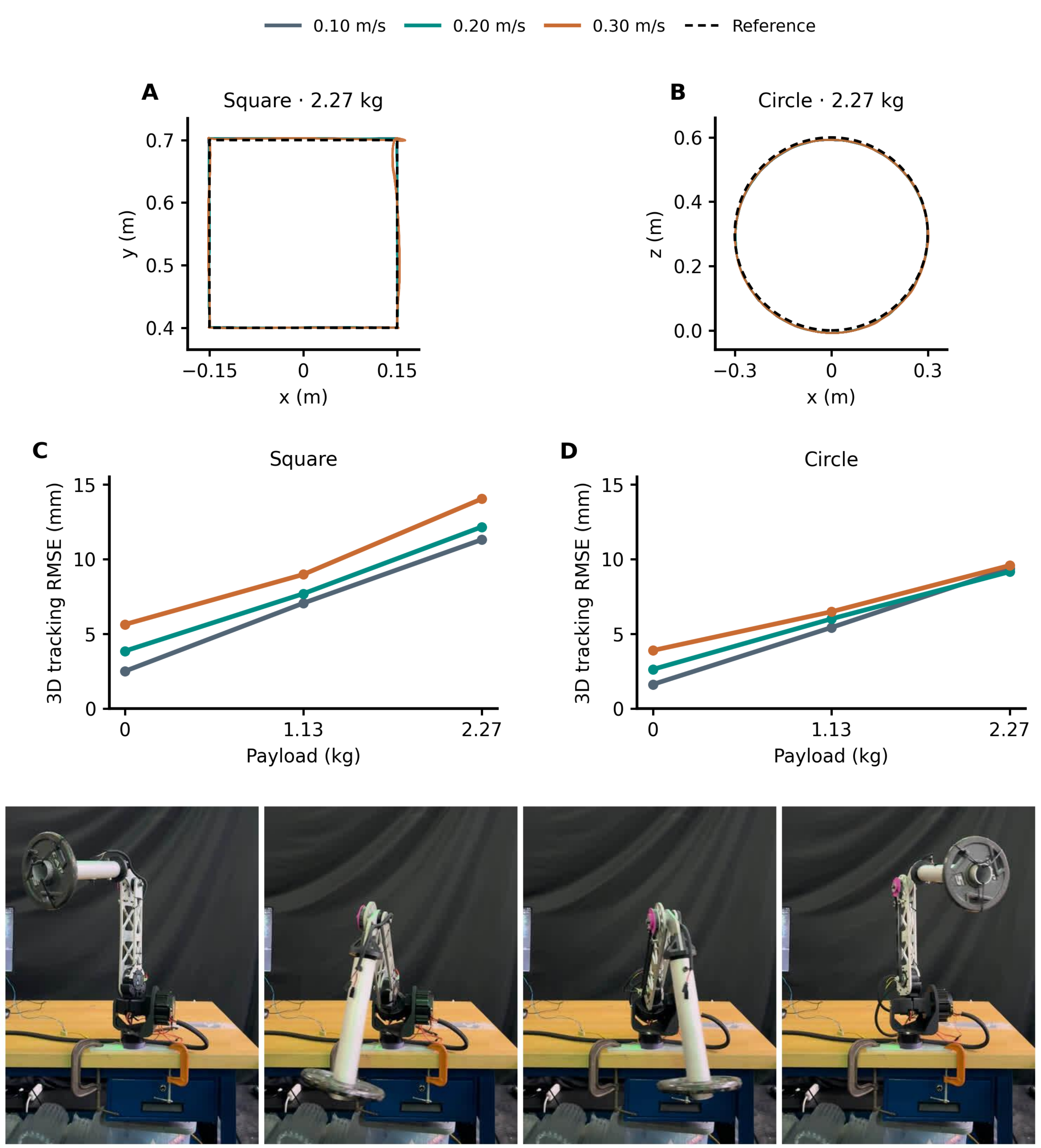}
    \caption{\textbf{SRL tracking under different payloads.} \textbf{(A, B)} Encoder-derived paths with a $5$ lb payload, projected onto the $xy$ plane for the square ($z = 0.30$ m) and the $xz$ plane for the circle ($y = 0.50$ m). The square has $0.30$ m sides, and the circle has a $0.30$ m radius; black dashed curves indicate commanded reference trajectories. Colors denote mean reference speeds of approximately $0.10$, $0.20$, and $0.30$ m/s. \textbf{(C, D)} Time-weighted 3D position RMSE versus payload for the square and circle trajectories, respectively. Each point summarizes approximately ten analyzed cycles from a single recording.}
    \label{fig:sp_payload_tracking}
\end{figure}

Fig.~\ref{fig:sr_exploded_view} shows the SRL assembly and its principal mechanical components. Table~\ref{tab:srl_system_specs} summarizes the SRL system specifications, Table~\ref{tab:joint_spec} provides joint and actuator specifications, and Table~\ref{tab:meche_comp} describes the principal mechanical components and fabrication details. Fig.~\ref{fig:sp_payload_tracking} shows the SRL tracking performance across different payloads and reference speeds for square and circular trajectories.

\begin{table}[!htbp]
\small
\centering
\caption{\textbf{SRL system and integration specifications}}
\begin{tabular}{l|cc}
\hline
Property & Specification \\ \hline
Degrees of freedom / arm length & $4$ / $83.0$ cm \\
Upper arm / forearm length & $37.3$ / $45.7$ cm \\
Total mounted mass & $4.7$ kg\\
Continuous / peak payload & $2$ / $4$ kg\\
Gripper mass & $0.5$ kg\\ 
Battery / approximate runtime & $36$ V, $8$ Ah / $\sim 65$ min\\
Power and joint communication & Daisy chain; CAN\\
Torso / SRL cameras & Intel RealSense D455 / ZED2i; RGB-D \\
\hline
\end{tabular}
\label{tab:srl_system_specs}
\par\smallskip
\begin{minipage}{0.73\linewidth}
\footnotesize
Arm length is the sum of the J3--J4 and J4--end-effector
distances, excluding the upstream mounting assembly.
\end{minipage}
\end{table}

\begin{table}[!htbp]
\small
\centering
\setlength{\tabcolsep}{3pt}
\caption{\textbf{Joint and actuator specifications of the SRL}}
\begin{tabular}{l|cccccc}
\hline
Joint &
Motion &
Actuator &
\shortstack{External ratio} &
\shortstack{Continuous torque} &
\shortstack{Peak torque} &
\shortstack{Joint range} \\ \hline

J1 & Shoulder yaw   & RS-02 & $1{:}1$ &
$\pm 6$ Nm & $\pm 17$ Nm & $[-180^\circ,180^\circ]$ \\

J2 & Shoulder pitch & RS-03 & $1{:}1$ &
$\pm 20$  Nm & $\pm 60$ Nm & $[-100^\circ,100^\circ]$\\

J3 & Shoulder roll  & RS-02 & $1{:}1$ &
$\pm 6$ Nm & $\pm 17$ Nm & $[-90^\circ,90^\circ]$\\

J4 & Elbow          & RS-02 & $3{:}1$ &
$\pm 6$ Nm & $\pm 17$ Nm & $[-125^\circ,50^\circ]$\\ \hline
\end{tabular}
\label{tab:joint_spec}

\par\smallskip
\begin{minipage}{0.90\linewidth}
\footnotesize
The RS-02 actuator has a $7.75{:}1$ internal reduction and a mass
of $405$\,g; the RS-03 has a $9{:}1$ internal reduction and a mass
of $880$\,g. External ratios refer to the transmission between the actuator output
and the joint. Listed torque ratings are measured at the actuator output,
before the external transmission. J4 uses a belt transmission.
\end{minipage}
\end{table}

\begin{table}[!htbp]
\small
\centering
\setlength{\tabcolsep}{4pt}
\caption{\textbf{Mechanical components and fabrication of the SRL}}
\begin{tabular}{l|ccc}
\hline
Component & Material & Fabrication & Detail \\ \hline

Main shoulder bracket &
PLA &
3D printed &
Actuator housing \\

Primary elbow plate &
AL-6061 &
Waterjet &
Elbow structure \\

Secondary elbow plate &
AL-6061 &
Waterjet &
Elbow structure \\

Elbow shells &
PLA &
3D printed &
Elbow enclosure \\

Forearm tube &
AL-6061 &
OTS &
Internal cable routing \\

Wrist shaft &
AL-7075 &
Machined &
Keyed interface \\

Neck-mount adapter &
PLA &
3D printed &
Neck attachment \\

Back-mount adapter &
PLA &
3D printed &
Back attachment \\ \hline
\end{tabular}
\label{tab:meche_comp}
\end{table}

\FloatBarrier
\subsection*{Workspace and Contact Analysis}
\subsubsection*{Workspace and manipulability} 
\begin{figure}[!htbp]
    \centering
    \includegraphics[width=1\linewidth]{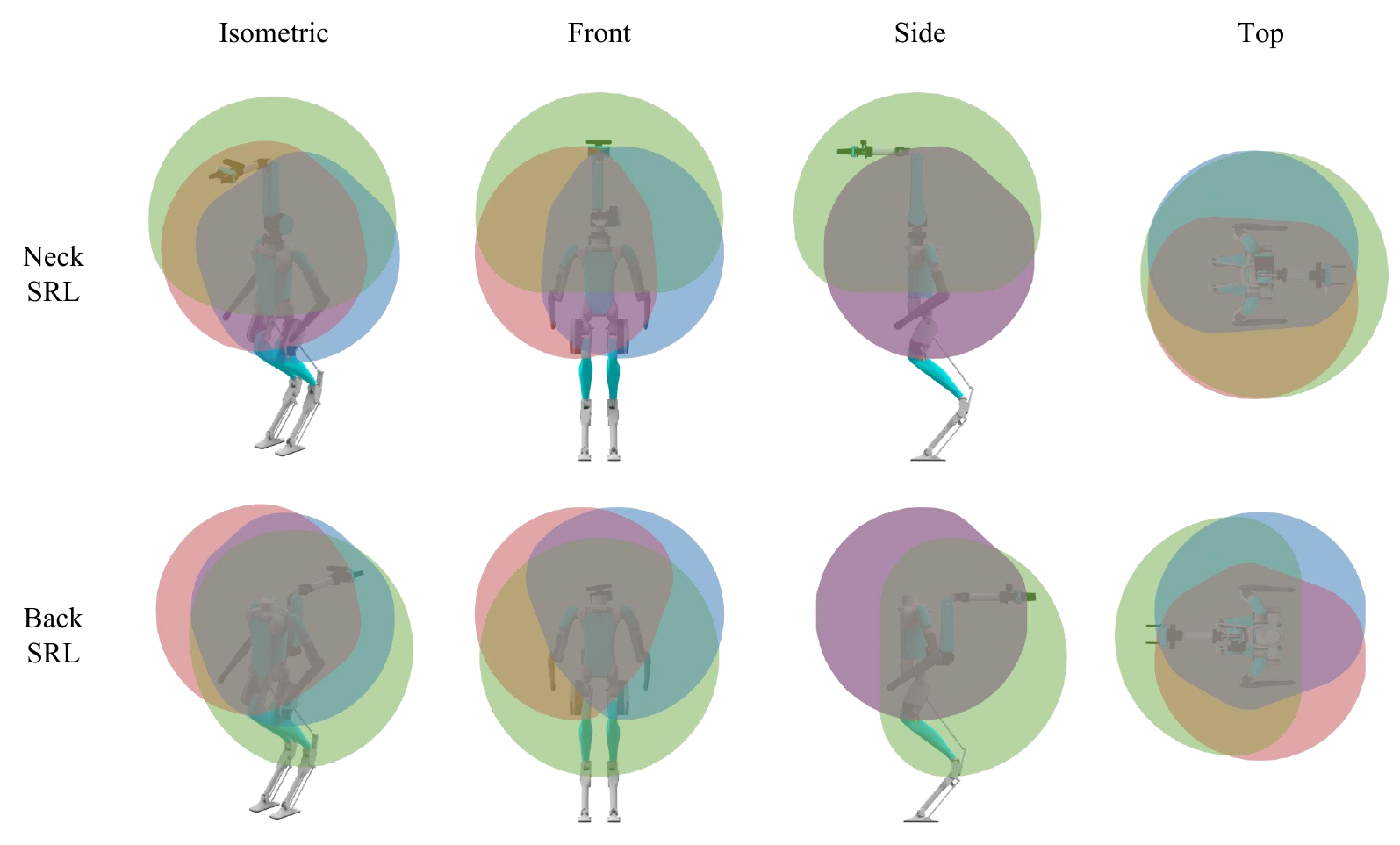}
    \caption{\textbf{Multi-view visualization of the augmented workspace.}
    Reachable workspaces for the neck-mounted (top row) and back-mounted (bottom row) configurations, displayed from isometric, front, side, and top views. Translucent volumes represent the reachable workspaces of the native arms and the SRLs, illustrating their spatial overlap.}
    \label{fig:sp_workspace_views}
\end{figure}

The workspace illustrations and volume summary in Fig.~\ref{fig:workspace}B use the original analysis, which specifies $5,000,000$ random joint configurations per arm and a $20$ mm voxel grid. Arms were sampled within joint limits with the torso and lower body in a nominal standing pose; configurations with detected self-collisions were rejected, and environmental obstacles were excluded. Occupied voxels were classified by single-arm reachability: ``exactly-one", ``exactly-two", and ``exactly-three" arm volumes sum to the total workspace union, while the ``at-least-two" volume combines the latter two. These overlaps classifications describe reachability by individual arms and do not guarantee collision-free simultaneous multi-arm configurations. Additional views are provided in Fig.~\ref{fig:sp_workspace_views}.

The heatmaps use the velocity-scaled translational manipulability isotropy (defined in Materials and Methods), evaluated using the full base-to-end-effector Jacobian at the nominal base pose. Each voxel records the maximum score sampled across all accessible arms. Zero-padding and Gaussian smoothing were applied solely for visual presentation and were excluded from the underlying distance-analysis workspace.

\subsubsection*{Workspace distance and contact metrics}
For Fig.~\ref{fig:independent_aug}E-H, signed distance is measured in the instantaneous body frame from the SRL reference to the native-arm voxel-union boundary (positive outside, negative inside). The workspace union is discretized into $20$ mm voxels using $2^{22}$ Sobol samples per arm, with shaded bands spanning $15$, $20$, and $25$ mm voxel estimates to illustrate resolution sensitivity.  Wall-support boundaries satisfy static force and moment equilibrium under fixed, unilateral contacts with a friction coefficient of $\mu=1$ for both  wall and floor, a $16$-ray friction cone linearization, and maximum total wall-normal force capped at $5/10/20\%$ of the modeled weight (shared between the two native hands). For Fig.~\ref{fig:coordinated_aug}F, the Ferrari–
Canny-type score $\epsilon$ (45) is the radius of the largest origin-centered ball contained within the convex hull of unit-normalized contact wrenches; $\epsilon=0$ if the origin lies outside or on the hull boundary.
Torque components were scaled by the modeled stack’s bounding-box diagonal, \(\ell=1.75\) m, before unit-wrench normalization. Four ray directions approximate the friction cone at each contact of the idealized rigid stack. Note that this normalized geometric metric evaluates wrench-space coverage rather than physical force capacity.

\newpage
\subsection*{Trajectory optimization implementation}
\label{sec:to_implementation}
For the neck-mounted TO example, the configuration vector contains $n_q=33$ coordinates: seven floating-base coordinates representing position and unit-quaternion orientation, together with $26$ internal joint coordinates. The generalized velocity and control dimensions are $n_v=32$ and $n_u=26$, respectively, giving a stored state vector $\mathbf{x}=[\mathbf{q};\mathbf{v}]$ with $65$ entries. The TO control dimension counts the joint-torque inputs in the optimization model, whereas the deployed RL controller outputs $24$ joint-position commands. Each active rigid foot contact contributes three force and three moment components to the stacked contact wrench. Table~\ref{tab:to_settings} summarizes the numerical settings.

\begin{table}[!htbp]
\centering
\small
\caption{\textbf{Trajectory-optimization settings for the neck-mounted example}}
\label{tab:to_settings}
\begin{tabular}{
    >{\raggedright\arraybackslash}p{.36\linewidth}
    >{\raggedright\arraybackslash}p{.56\linewidth}
}
\toprule
Setting & Value or implementation \\
\midrule
Solver
& Crocoddyl FDDP \\

Time integration
& Symplectic (semi-implicit) Euler \\

Time step
& $\Delta t=0.005$ s \\

Control parameterization
& Piecewise-constant joint torques \\

Maximum iterations
& $200$ per phase \\

Solver stopping threshold
& $10^{-8}$ \\

State initialization
& Phase-initial state repeated across trajectory nodes \\

Control initialization
& Quasi-static joint torques \\

Foot-contact model
& Six-dimensional rigid contacts \\

Foot-contact friction coefficient
& $\mu_{\mathrm{TO}}=1$ \\

Standing friction approximation
& Four-facet outer friction-pyramid approximation \\
\bottomrule
\end{tabular}
\end{table}

\newpage
\subsection*{RL Policy Training Details}
\label{app:training_detail}
\subsubsection*{Observation space}
The complete student and teacher observation spaces are summarized in Table~\ref{tab:obs_space}. During training, observation noise is added to the student inputs to model onboard sensing uncertainty, while the teacher receives noise-free ground truth from simulation.

\begin{table}[!htbp]
    \centering
    \small
    \caption{Observation spaces.}
    \begin{tabular}{lcccc}
        \toprule
        \textbf{Input} & \textbf{Symbol} & \textbf{Dimension} & \textbf{Student} & \textbf{Teacher} \\
        \midrule
        Base Linear Velocity           & $\mathbf{\dot{p}}_{t}^{\rm b}$        & $3$ & \checkmark & \checkmark \\
        Base Angular Velocity           & $\omega^{\rm b}_{t}$         & $3$ & \checkmark & \checkmark \\
        Projected Gravity        & $\mathbf{g}_{t}$       & $3$ & \checkmark & \checkmark \\
        History Motor Joint Position   & $\mathbf{q}^{\rm j}_{\rm hist}$     & $200$ & \checkmark & \checkmark \\
        Motor Joint Velocity         & $\mathbf{\dot{q}}^{\rm j}_{t}$              & $20$ & \checkmark & \checkmark \\
        Ref. End-effector Position    & $\hat{\mathbf{{p}}}^{\rm ee}_{t}$    & $15$  & \checkmark  & \checkmark \\
        History Action            & $\mathbf{a}_{\rm hist}$             & $200$ & \checkmark & \checkmark \\
 
        \midrule
        Base Translation         & $\mathbf{p}^{\rm b}_t$        & $3$   &   & \checkmark \\
        Base Orientation         & $\boldsymbol{\theta}^{b}_t$        & $3$   &   & \checkmark \\
        End-effector Position        & $\mathbf{p}^{\rm e}_t$      & $12$  &   & \checkmark \\
        Ref. Base Translation     & $\mathbf{\hat{p}}^{\rm b}_t$  & $3$   &   & \checkmark \\
        Ref. Base Orientation     & $\hat{\boldsymbol{\theta}}^{b}_t$  & $3$   &   & \checkmark \\
        Ref. Base Linear Velocity       & $\mathbf{\hat{\dot{p}}}^{\rm b}_t$  & $3$ &  & \checkmark \\
        Ref. Base Angular Velocity       & $\mathbf{\hat{\omega}}^{\rm b}_t$  & $3$ &  & \checkmark \\
        Ref. Actuator Joint Position     & $\mathbf{\hat{q}}^{\rm j}_t$     & $20$ &  & \checkmark \\
        PD Gains                 & $\mathbf{K}_p, \mathbf{K}_d$       & $40$  &   & \checkmark\\
        Motor Strength Scale     & $\boldsymbol{\alpha}$  & $20$  &   & \checkmark \\
        
        \bottomrule
\end{tabular}
\label{tab:obs_space}
\end{table}

\FloatBarrier
\subsubsection*{Action space}
The policy action $\mathbf{a}_t$ specifies joint-position offsets relative to a default standing configuration $\mathbf{q}^{\mathrm{j}}_{\mathrm{init}}$ for the $24$ actuated joints. The target joint positions are converted to torque commands via a PD controller:
$\boldsymbol{\tau}_t = \mathbf{K}_p (\mathbf{q}^{\mathrm{j}}_{\mathrm{init}} + \mathbf{a}_t - \mathbf{q}^{\mathrm{j}}_t) - \mathbf{K}_d \dot{\mathbf{q}}^{\mathrm{j}}_t$,
where $\mathbf{K}_p$ and $\mathbf{K}_d$ are the joint-level proportional and derivative gains, respectively. The control loop operates at $200$~Hz, while policy targets are updated at $1$~kHz during training and $2$~kHz during deployment. No additional filtering is applied to the policy outputs.

\FloatBarrier
\subsubsection*{Reward deign}
The reward function encourages the policy to track the trajectory optimization (TO)-generated reference motion while maintaining stable and physically feasible whole-body behavior. It balances task-tracking objectives with regularization terms that penalize undesirable joint motions, excessive control effort, and task-dependent constraint violations. The individual reward terms and their corresponding weights are summarized in Table~\ref{tab:reward}.

\begin{table}[!htbp]
\caption{Reward Components and Weights.}
\centering
\small
\label{tab:reward}
\begin{tabular}{l l l c}
\midrule
\textbf{Category} 
& \textbf{Term} 
& \textbf{Expression} 
& \textbf{Weight} \\
\midrule
\multirow{8}{*}{\makecell{Task \\ Reward}}
&Motor Joint Position 
& $\exp(-5 \|\mathbf{{\hat{q}}}^{\rm j}_t - \mathbf{{{q}}}^{\rm j}_t\|_2^2)$                 
& $3$ \\
& Base Position 
& $\exp(-20 \|\mathbf{\hat{p}}^{\rm b}_t - \mathbf{{p}}^{\rm b}_t\|_2^2)$                 
& $3$ \\
& Base Orientation 
& $\exp(-50 \|\hat{\boldsymbol{\theta}}^{\rm b}_t - {\boldsymbol{\theta}}^{\rm b}_t\|_2^2)$       
& $3$ \\
& Base Linear Velocity 
& $\exp(-2 \| \mathbf{\hat{\dot{p}}}_t^{\rm b} - \mathbf{{\dot{p}}}^{\rm b}_t\|_2^2)$             
& $3$ \\
& Base Angular Velocity 
& $\exp(-0.5 \|\hat{\boldsymbol{\omega}}^{\rm b}_t - {\boldsymbol{\omega}}^{\rm b}_t\|_2^2)$ 
& $3$ \\
& End-effector Position 
& $\exp(-20 \|\hat{\mathbf{{p}}}^{\rm ee}_t - {\mathbf{{p}}}^{\rm ee}_t\|_2^2)$         
& $3$ \\
\midrule
\multirow{3}{*}{\makecell{Penalty\\ Cost}} 
& Action Rate 
& $\|\mathbf{a}_t - 2\mathbf{a}_{t-1} + \mathbf{a}_{t-2}\|_2^2$            
& $-3$ \\
& Torques 
& $\|\mathbf{u}_t / \mathbf{u}_{\text{limit}}\|_2^2$                                        
& $-0.25$ \\
& Joint Acceleration 
& $\|\mathbf{{\ddot{q}}}^{\rm j}_t\|_2^2$                                        
& $-10^{-5}$ \\
\midrule
\end{tabular}
\end{table}

\FloatBarrier
\subsubsection*{Domain randomization and curriculum}
To mitigate the sim-to-real gap and facilitate zero-shot hardware transfer, physical and sensing parameters are randomized during training. The randomized quantities encompass robot dynamics, contact properties, observation perturbations, and task-dependent parameters. Unless otherwise specified, these parameters are sampled independently at the beginning of each episode from the ranges listed in Table~\ref{tab:domain_randomization}.

\begin{table}[!htbp]
\caption{Domain Randomization Parameters.}
\centering
\small
\label{tab:domain_randomization}
\begin{tabular}{l l l l}
\toprule
\textbf{Category} & \textbf{Parameter} & \textbf{Type} & \textbf{Range / Std} \\
\midrule
\multirow{6}{*}{Observation} 
& Ref. End-effector Position (m) 
& Additive Gaussian 
& $\sigma=0.04$ \\
& Motor Joint Position (rad) 
& Additive Gaussian 
& $\sigma=0.035$ \\
& Joint Velocity (rad/s)
& Additive Gaussian 
& $\sigma=0.03$ \\
& Base Linear Velocity (m/s)
& Additive Gaussian 
& $\sigma=0.03$ \\
& Base Angular Velocity (rad/s)
& Additive Gaussian 
& $\sigma=0.03$ \\
& Projected gravity direction (unitless) 
& Gaussian; unit norm 
& $\sigma=0.015$ \\
\midrule
Delays 
& Action Delay (ms)
& Discrete uniform 
& $\{0, 5, 10, 15, 20\}$ \\
\midrule
\multirow{2}{*}{Motor} 
& Motor Strength 
& Uniform scaling 
& $[0.95, 1.05]$ \\
& Kp/Kd Factor 
& Uniform scaling 
& $[0.9, 1.1]$ \\
\midrule
\multirow{3}{*}{Reset} 
& Initial joint position (rad) & Additive Gaussian & $\sigma=0.03$ \\
& Initial joint velocity (rad/s) & Zero-mean Gaussian & $\sigma=0.06$ \\
& Initial base linear / angular velocity & Zero-mean Gaussian & $\sigma=0.05$ m/s / $0.05$ rad/s  \\
\midrule
Body 
& Mass 
& Uniform scaling 
& $[0.9, 1.1]$ \\
\midrule
\multirow{1}{*}{Environment} 
& Gravity 
& Uniform scaling 
& $[0.9, 1.1]$ \\
& Friction 
& Uniform scaling 
& $[0.5, 1.0]$ \\
\bottomrule
\end{tabular}
\end{table}

\subsubsection*{Training hyperparameters}
The teacher and student policies were trained using the L2T framework described in the main text. The principal optimization and network hyperparameters are summarized in Table~\ref{tab:rl_training_hyperparameters}.

\begin{table}[!htbp]
\small
\centering
\caption{\textbf{RL policy training hyperparameters.}}
\begin{tabular}{l|c}
\hline
Parameter & Value \\ \hline
PPO clipping parameter $\epsilon$ & $0.3$ \\
Discount factor $\gamma$ & $0.97$ \\
Learning rate & $5\times10^{-5}$ \\
Entropy coefficient & $1\times10^{-2}$ \\
Rollout length & $20$ \\
Number of minibatches & $32$ \\
Updates per batch & $4$ \\
Number of parallel environments & $4096$ \\
Batch size & $128$ \\
Maximum gradient norm & $1.0$ \\
Observation normalization & Yes \\
Teacher network & $(512,512,256,256)$ \\
Student network & $(512,512,256,256)$ \\ \hline
\end{tabular}
\label{tab:rl_training_hyperparameters}
\end{table}

\newpage
\subsubsection*{Controller evaluation}
\begin{figure}[!htbp]
    \centering
    \includegraphics[width=0.7\linewidth]{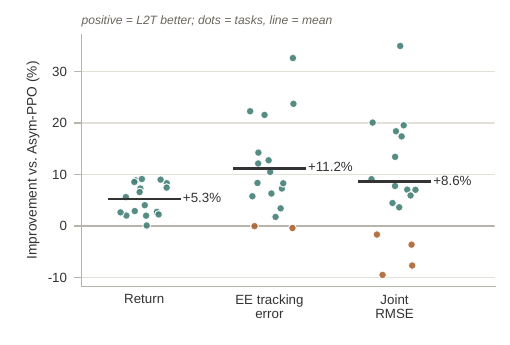}
    \caption{\textbf{Distribution of controller performance improvements.} Relative improvements of the L2T student over Asym-PPO in episodic return, end-effector tracking, and joint-position tracking. Each dot represents an evaluated task, and horizontal bars indicate averages across tasks. Positive values indicate higher return or lower tracking error for L2T; negative values favor Asym-PPO.}
    \label{fig:l2t-summary}
\end{figure}

The L2T student achieves higher average normalized return and lower average end-effector and joint-position tracking errors than Asym-PPO (Table~\ref{tab:l2t-summary}). Improvements vary across tasks, with some tasks exhibiting increased tracking error (Fig.~\ref{fig:l2t-summary}).

\begin{table}[!htbp]
\small
\centering
\caption{\textbf{Comparison of whole-body controller performance.}
Both policies use onboard observations. Values are task averages, with episodic return normalized by the privileged teacher for each task. Relative improvements are computed per task and then averaged; positive values
indicate higher return or lower tracking error. MPKPE measures end-effector position error in the robot's base frame. Results use one training seed per task.}
\label{tab:l2t-summary}
\begin{tabular}{@{}l r r r r@{}}
\toprule
& & & \multicolumn{2}{c}{$\Delta$ vs.\ Asym-PPO (\%)} \\
\cmidrule(l){4-5}
Metric & Asym-PPO & L2T (ours) & Mean & Max \\
\midrule
Episodic return (teacher $=1$) $\uparrow$ & 0.916 & \best{0.964} & \best{+5.3} & +9.1 \\
End-effector MPKPE (mm) $\downarrow$ & 124.1 & \best{113.3} & \best{+11.2} & +32.6 \\
Joint-position RMSE (rad) $\downarrow$ & 0.053 & \best{0.048} & \best{+8.6} & +35.0 \\
\bottomrule
\end{tabular}
\end{table}

\subsection*{Vision-guided autonomous control implementation}
\subsubsection*{Data collection and synchronization}
Demonstrations were collected on hardware using trained task-specific RL student controllers to execute reference motions in predefined scenes. RGB-D streams from the torso-mounted and SRL end-effector cameras were recorded at $15$ Hz, while proprioceptive states and reference commands were recorded at $200$ Hz. Each control sample was paired with the latest available image from each camera. The synchronized trajectories were subsampled to $50$ Hz for diffusion-policy training, retaining only successful runs.

\subsubsection*{Observation and action specification}
The diffusion policy uses a single observation and predicts an action sequence of length $l_p=64$, executing the first $l_a=32$ actions before the next inference. The five controlled end-effectors are the two native hands, two feet, and the SRL end-effector, with their reference positions expressed in the robot body frame. The policy-mode variable enables transitions between task-specific RL controllers when required by the task. Table~\ref{tab:dp_obs_action} summarizes the implementation details.

\subsubsection*{Diffusion-policy implementation and online inference}
We implement the high-level planner using the U-Net diffusion policy architecture in LeRobot \cite{cadene2024lerobot}. Visual observations are encoded with a ResNet-18 backbone adapted for RGB-D input and pre-trained with R3M weights, while FiLM conditions the network on observation features and diffusion timesteps. Training follows the conditional DDPM objective outlined in the main text, with the network parameterized to predict the clean action chunk, $\hat{A}_t = D_\theta(A_t^k, O_t, k)$, rather than the noise. The corresponding noise estimate is $\boldsymbol{\xi}_\theta = (A_t^k - \sqrt{\bar{\alpha}_k}\,\hat{A}_t)/\sqrt{1-\bar{\alpha}_k}$, so that
\begin{equation}
\|A_t - \hat{A}_t\|_2^2 = \frac{1-\bar{\alpha}_k}{\bar{\alpha}_k}\,\|\boldsymbol{\xi} - \boldsymbol{\xi}_\theta(A_t^k, O_t, k)\|_2^2 .
\end{equation}
We minimize the unweighted reconstruction error $\|A_t - \hat{A}_t\|_2^2$, which corresponds to the main-text objective with $w(k) = (1-\bar{\alpha}_k)/\bar{\alpha}_k$, i.e., the inverse signal-to-noise ratio at step $k$. This weighting emphasizes high-noise steps, where the policy must generate a coherent action chunk from nearly pure noise. We train a separate diffusion policy for each autonomous task; key network and hyperparameter details are summarized in Table~\ref{tab:dp_hyperparameters}.

At deployment, the diffusion policy runs in a receding-horizon fashion, generating short-horizon reference actions from updated visual and proprioceptive state inputs. These predicted action chunks are temporally interpolated to match the higher-frequency RL control loop before being sent to the task-specific student controller. 
The RL controllers handle lower-level whole-body stabilization and trajectory tracking, while the diffusion policy generates high-level references, gripper commands, and controller selection signals.

\begin{table}[!htbp]
\small
\centering
\caption{\textbf{Diffusion-policy observation and action specifications.}
Sensor rates are logging rates; training data are resampled to $50$ Hz,
with each sample paired with the most recent images.
Check marks indicate signals used by each task-specific policy:
top-shelf retrieval (Shelf), stacked-box stabilization (Box),
and object transport during door interaction (Door).
}
\setlength{\tabcolsep}{3pt}
\begin{tabular}{l|lll|ccc}
\hline
Type & Signal & Dim. & Rate & Shelf & Box & Door \\ \hline
Observation & Torso / SRL end-effector RGB-D & $240\times424\times4$ each & 15 Hz & \checkmark & \checkmark & \checkmark \\
& Motor positions & $24$ & 200 Hz & \checkmark & \checkmark & \checkmark \\
& Gripper state (last command) & $1$ & 200 Hz & \checkmark & & \\
& Policy mode (last command) & $1$ & 200 Hz & & & \checkmark \\
Action & End-effector positions (body frame) & $3\times5$ & 50 Hz & \checkmark & \checkmark & \checkmark \\
& Base pose (body frame) & $6$ & 50 Hz & \checkmark & \checkmark & \\
& Gripper command & $1$ & 50 Hz & \checkmark & & \\
& Policy mode (RL controller index) & $1$ & 50 Hz & & & \checkmark \\ \hline
\multicolumn{4}{l|}{Total proprioceptive / action dimension} & $25/22$ & $24/21$ & $25/16$ \\ \hline
\end{tabular}
\label{tab:dp_obs_action}
\end{table}

\begin{table}[!htbp]
\small
\centering
\caption{\textbf{Diffusion-policy architecture and training parameters}
}
\begin{tabular}{l|ll}
\hline
Category & Parameter & Value \\ \hline

Architecture
& Diffusion backbone & 1D conv.\ U-Net (LeRobot) \\
& U-Net channels & $(512, 1024, 2048)$ (shelf, box); $(128,256,512)$ (door)\\
& Kernel size / GroupNorm groups & $5$ / $8$ \\
& Timestep embedding dim. & $128$ \\
& Conditioning & FiLM (scale and shift) \\
& Visual encoder & ResNet-18, GroupNorm, shared across cameras \\
& Encoder initialization & R3M; depth filters zero-initialized \\
& Image pooling & Spatial softmax ($32$ keypoints) \\
& Visual input & RGB-D (both cameras) \\ \hline

Training
& Dataset sampling rate & 50 Hz \\
& Observation / prediction horizon  & $l_o=1$ / $l_p=64$\\
& Diffusion  & $K=100$ steps; squared-cosine noise schedule\\
& Prediction target & Clean action chunk (sample) \\
& Loss weighting $w(k)$ & $(1-\bar{\alpha}_k)/\bar{\alpha}_k$ \\
& State/action normalization & Min--max to $[-1, 1]$ \\
& Optimizer & Adam, $\beta = (0.95, 0.999)$; weight decay $10^{-6}$ \\
& Learning rate & $10^{-4}$, cosine decay, $500$ warm-up steps \\
& Batch size & $8$ (shelf, box); $64$ (door) \\
& Training steps & $2\times10^{5}$ (shelf, door); $3\times10^{5}$ (box) \\
& Train/validation split & $19/3$ episodes (door); all episodes (shelf, box) \\ \hline

Inference
& Executed actions per chunk $l_a$ & $32$ \\
& Sampler & DDIM, $50$ denoising steps \\
& DP output / RL control rate & $50$ / $200$ Hz; linear action interpolation \\
& Chunk blending & Linear, $10$-step window\\ \hline

\end{tabular}
\label{tab:dp_hyperparameters}
\end{table}

\clearpage

\paragraph{Caption for Movie S1.}
\textbf{SRL embodiments and workspace exploration.} The video shows hardware demonstrations of workspace exploration with the neck-mounted and back-mounted SRLs, together with forward walking, backward walking, and turning. These demonstrations complement the morphological characterization in Fig.~\ref{fig:workspace}.

\paragraph{Caption for Movie S2.}
\textbf{Support and stability augmentation.} The video compares CAREN perturbation experiments with and without SRL support against the external frame, followed by wall-assisted leaning while the native arms hold a payload (Fig.~\ref{fig:force_aug}).

\paragraph{Caption for Movie S3.}
\textbf{Reach-augmented manipulation.} The video shows high-shelf retrieval and human delivery, front-to-rear shelf transfer, wall-supported reach-around retrieval, and rear manipulation concurrent with native-arm activity in front of the robot (Fig.~\ref{fig:independent_aug}).

\paragraph{Caption for Movie S4.}
\textbf{Coordinated multi-arm manipulation.} The video shows coordinated shelf relay, loading an object into a container held by the native arms, and stacked-box stabilization. These tasks illustrate how the three arms share acquisition, placement, and support roles (Fig.~\ref{fig:coordinated_aug}).

\paragraph{Caption for Movie S5.}
\textbf{Dynamic and collaborative whole-body manipulation.} The video shows collaborative table transport with a human partner, box carrying while the neck-mounted SRL pushes a door, and door-handle actuation with the back-mounted SRL while the native arms retain a carried box (Fig.~\ref{fig:dynamic_aug}).

\paragraph{Caption for Movie S6.}
\textbf{Vision-guided autonomous execution.} The video shows autonomous high-shelf retrieval and human delivery, stacked-box stabilization, and box transport during door interaction. Task-specific diffusion policies update reference targets from visual and proprioceptive observations, while the learned whole-body controllers execute the motions.

\end{document}